%% file: predmarketJMLR.tex
\documentclass[twoside,11pt]{article}

\usepackage{jmlr2e}
\usepackage{graphicx}
\usepackage{natbib}
\usepackage{amsmath}
\usepackage{amssymb}
\usepackage{floatflt}
\usepackage{subfigure}
\usepackage{algorithm}
\usepackage{algorithmic}
\usepackage{color}
\usepackage{hyperref}

\graphicspath{{figs/}{../pics/}{}}
\DeclareGraphicsExtensions{.eps}

\input{notation}

\def\argmin{\mathop{\rm argmin}}
\def\RR{\mathbb R}

\newtheorem{rem}[theorem]{Remark}
\newtheorem{asum}{Assumption}

\jmlrheading{-}{2012}{-}{-}{-}{Adrian Barbu and Nathan Lay}

\ShortHeadings{Artificial Prediction Markets}{Barbu and Lay}
\firstpageno{1}

\begin{document} 

\title{An Introduction to Artificial Prediction Markets for Classification}


\author{\name Adrian Barbu \email abarbu@fsu.edu \\
       \addr Department of Statistics\\
       Florida State University\\
       Tallahassee, FL 32306, USA
       \AND
       \name Nathan Lay \email nlay@fsu.edu \\
       \addr Department of Scientific Computing\\
       Florida State University\\
       Tallahassee, FL 32306, USA}

\editor{}

\maketitle

\begin{abstract}
Prediction markets are used in real life to predict outcomes of interest such as presidential elections. This paper presents a mathematical theory of artificial prediction markets for supervised learning of conditional probability estimators. The artificial prediction market is a novel method for fusing the prediction information of features or trained classifiers, where the fusion result is the contract price on the possible outcomes. The market can be trained online by updating the participants' budgets using training examples. Inspired by the real prediction markets, the equations that govern the market are derived from simple and reasonable assumptions. Efficient numerical algorithms are presented for solving these equations. The obtained artificial prediction market is shown to be a maximum likelihood estimator. It generalizes linear aggregation, existent in boosting and random forest, as well as logistic regression and some kernel methods. Furthermore, the market mechanism allows the aggregation of specialized classifiers that participate only on specific instances. Experimental comparisons show that the artificial prediction markets often outperform random forest and implicit online learning on synthetic data and real UCI datasets. Moreover, an extensive evaluation for pelvic and abdominal lymph node detection in CT data shows that the prediction market improves adaboost's detection rate from $79.6\%$ to $81.2\%$ at 3 false positives/volume.
\end{abstract}


%

\section{Introduction}

Prediction markets, also known as information markets, are forums that trade contracts that yield payments dependent on the outcome of future events of interest. They have been used in the US Department of Defense \citep{polk2003pam}, health care \citep{polgreen2006upm}, to predict presidential elections \citep{wolfers2004pm} and in large corporations to make informed decisions \citep{cowgill2008upm}. The prices of the contracts traded in these markets are good approximations for the probability of the outcome of interest \citep{manski2006interpreting,gjerstad2005risk}. prediction markets are capable of fusing the information that the market participants possess through the contract price. For more details, see \cite{arrow_promise_2008}.

In this paper we introduce a mathematical theory for simulating prediction markets numerically for the purpose of supervised learning of probability estimators. We derive the mathematical equations that govern the market and show how can they be solved numerically or in some cases even analytically. An important part of the prediction market is the contract price, which will be shown to be an estimator of the class-conditional probability given the evidence presented through a feature vector $\bf x$. It is the result of the fusion of the information possessed by the market participants. 

The obtained artificial prediction market turns out to have good modeling power. It will be shown in Section \ref{sec:ctbet} that it generalizes linear aggregation of classifiers, the basis of boosting \citep{friedman2000alr,schapire2003bam} and random forest \citep{breiman_random_2001}. It turns out that to obtain linear aggregation, each market participant purchases contracts for the class it predicts, regardless of the market price for that contract. Furthermore, in Sections \ref{sec:logbet} and \ref{sec:svm} will be presented special {\em betting functions} that make the prediction market equivalent to a logistic regression and a kernel-based classifier respectively.

 We introduce a new type of classifier that is {\em specialized} in modeling certain regions of the feature space. Such classifiers have good accuracy in their region of specialization and are not used in predicting outcomes for observations outside this region. This means that for each observation, a different subset of classifiers will be aggregated to obtain the estimated probability, making the whole approach become a sort of {\em ad-hoc} aggregation. This is contrast to the general trend in boosting where the same classifiers are aggregated for all observations.

We give examples of generic specialized classifiers as the leaves of random trees from a random forest. Experimental validation on thousands of synthetic datasets with Bayes errors ranging from 0 (very easy) to 0.5 (very difficult) as well as on real UCI data show that the prediction market using the specialized classifiers outperforms the random forest in prediction and in estimating the true underlying probability.

Moreover, we present experimental comparisons on many UCI datasets of the artificial prediction market with the recently introduced implicit online learning \citep{kulis2010implicit} and observe that the market significantly outperforms the implicit online learning on some of the datasets and is never outperformed by it.

\section{The Artificial Prediction Market for Classification}

This work simulates the {\em Iowa electronic market} \citep{wolfers2004pm}, which is a real prediction market that can be found online at http://www.biz.uiowa.edu/iem/. 

\subsection{The Iowa Electronic Market}

The {\em Iowa electronic market} \citep{wolfers2004pm} is a forum where contracts for future outcomes of interest (e.g. presidential elections) are traded.

Contracts are sold for each of the possible outcomes of the event of interest. The contract price fluctuates based on supply and demand. In the Iowa electronic market, a winning contract (that predicted the correct outcome) pays \$1 after the outcome is known. Therefore, the contract price will always be between 0 and 1.

Our market will simulate this behavior, with contracts for all the possible outcomes, paying 1 if that outcome is realized.

\subsection{Setup of the Artificial Prediction Market}\label{sec:setup}

If the possible classes (outcomes) are $1,...,K$, we assume there exist contracts for each class, whose prices form a $K$-dimensional vector ${\bf c}=(c_1,...,c_K)\in \Delta \subset [0,1]^K$, where $\Delta$ is the probability simplex $\Delta =\{ {\bf c}\in [0,1]^K, \sum_{k=1}^K c_k=1\}$.

Let $\Omega\subset \RR^F$ be the instance or feature space containing all the available information that can be used in making outcome predictions $p(Y=k|{\bf x}),{\bf x}\in \Omega$. 

The market consists of a number of market participants $(\beta_m,\phi_m({\bf x},{\bf c})), m=1,...,M$. 

A {\em market participant} is a pair $(\beta,\phi({\bf x},{\bf c}))$ of a {\em budget} $\beta$ and a {\em betting function} $\phi({\bf x},{\bf c}):\Omega\times \Delta \to [0,1]^K, \phi({\bf x},{\bf c})=\left( \phi^1({\bf x},{\bf c}),..., \phi^K({\bf x},{\bf c})\right) $. The budget $\beta$ represents the weight or importance of the participant in the market. 
 The betting function tells what percentage of its budget this participant will allocate to purchase contracts for each class, based on the instance ${\bf x}\in \Omega$ and the market price $\bf c$. As the market price $\bf c$ is not known in advance, the betting function describes what the participant plans to do for each possible price $\bf c$. The betting functions could be based on trained classifiers $h({\bf x}):\Omega\to \Delta, h({\bf x})=( h^1({\bf x}),..., h^K({\bf x})), \sum_{k=1}^K h^k({\bf x})=1$, but they can also be related to the feature space in other ways. We will show that logistic regression and kernel methods can also be represented using the artificial prediction market and specific types of betting functions. In order to bet at most the budget $\beta$, the betting functions must satisfy $\sum_{k=1}^K \phi^k({\bf x},{\bf c}))\leq 1$.
 
 \begin{figure*}[htb]
\centering
\hspace{-1mm}\includegraphics[height=2.8cm]{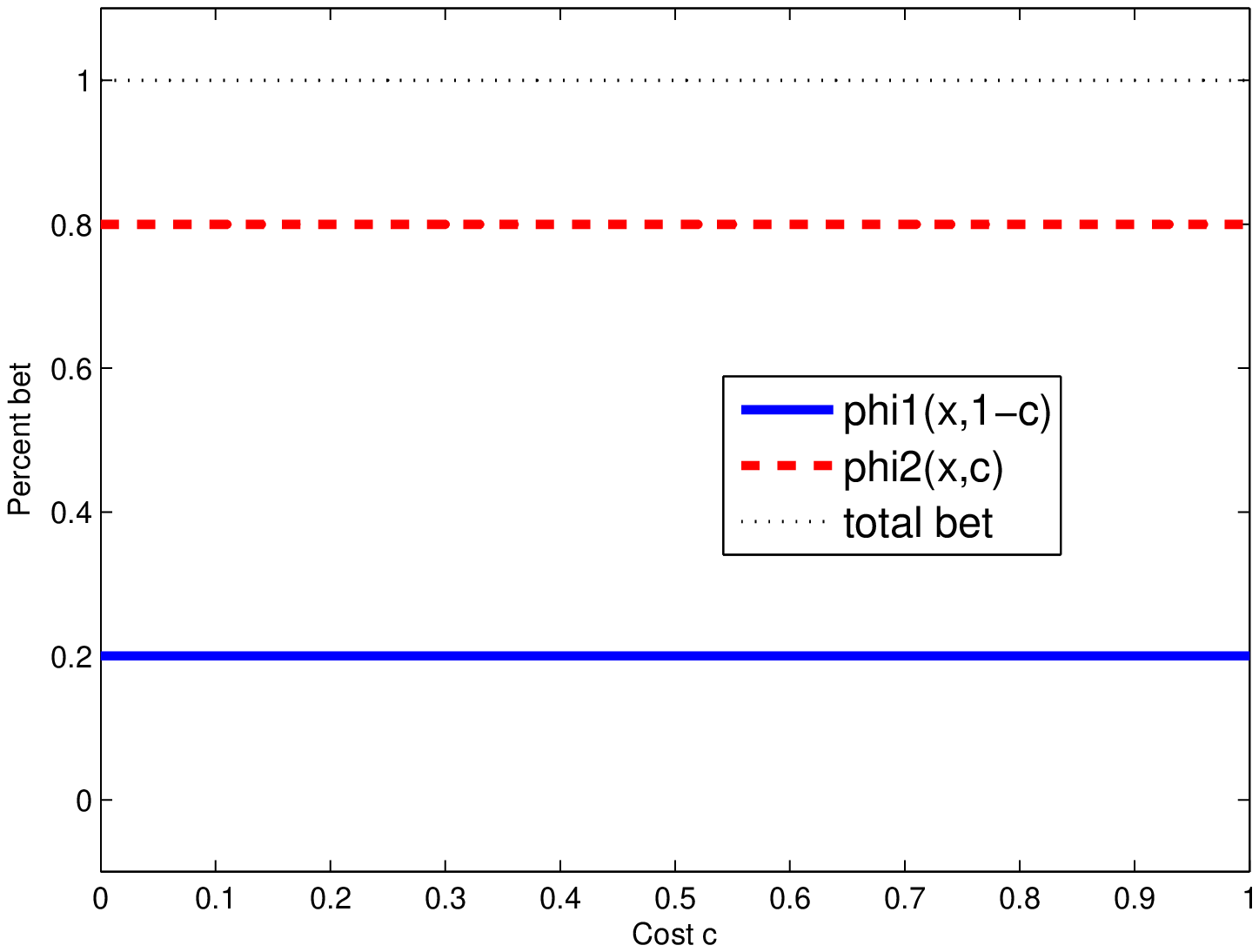}
\hspace{-1mm}\includegraphics[height=2.8cm]{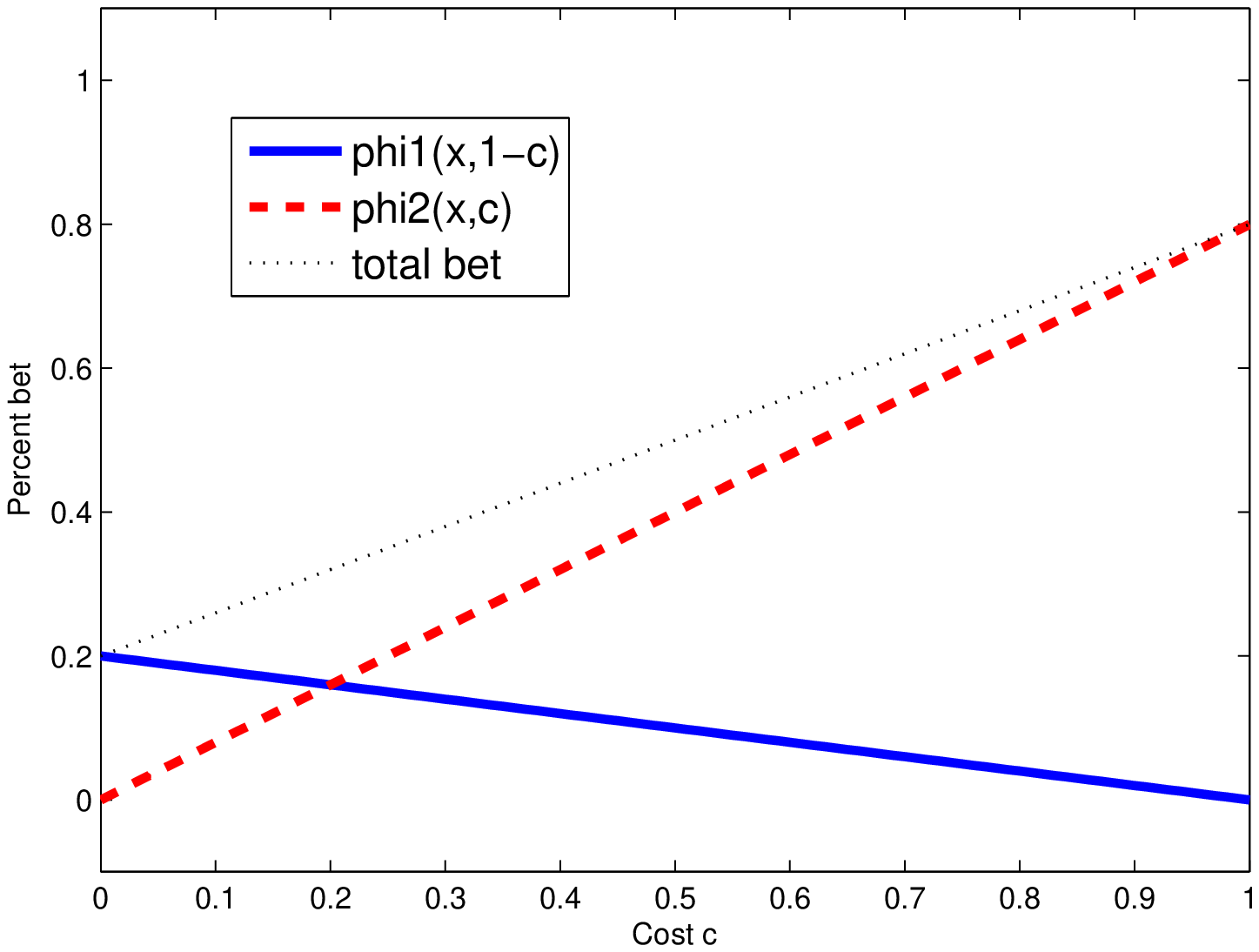}
\hspace{-1mm}\includegraphics[height=2.8cm]{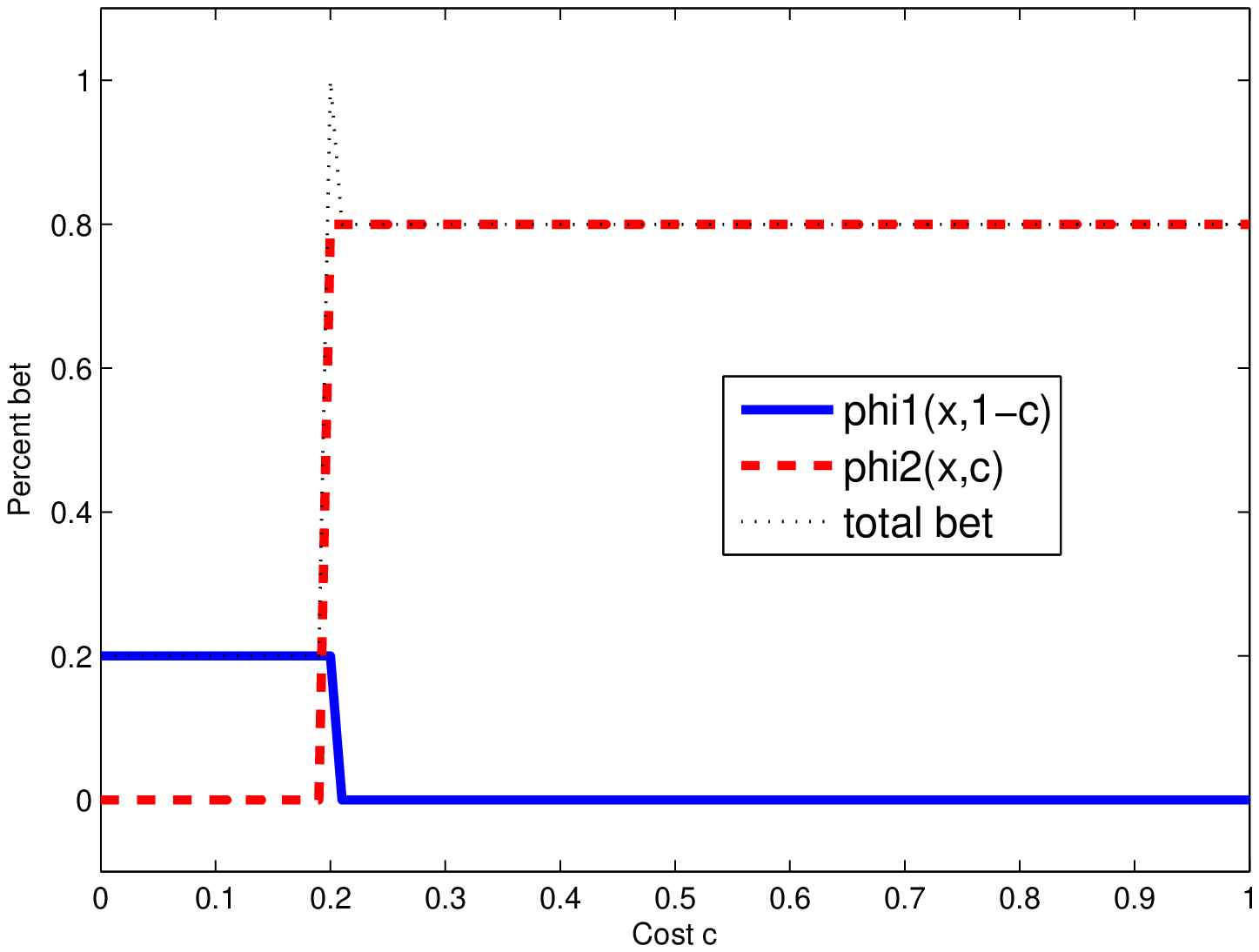}
\hspace{-1mm}\includegraphics[height=2.8cm]{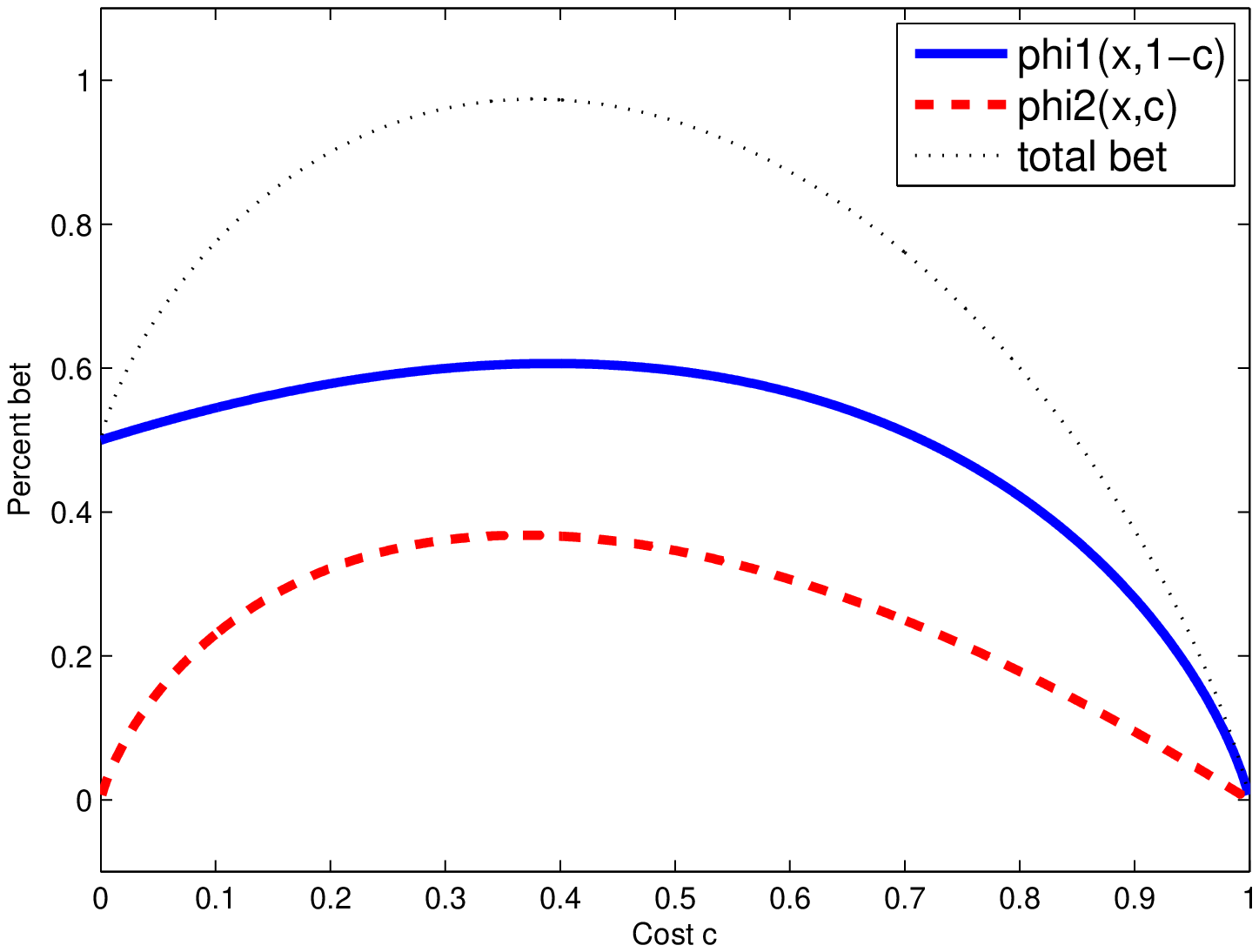}
\vskip -3mm
\caption{ Betting function examples: a) Constant, b) Linear, c) Aggressive, d) Logistic. Shown are $\phi^1({\bf x},1-c)$ (red), $\phi^2({\bf x},c)$ (blue), and the total amount bet $\phi^1({\bf x},1-c)+\phi^2({\bf x},c)$ (black dotted). For a) through c), the classifier probability is $h^2({\bf x})=0.2$.}
\label{fig:betfun}
\end{figure*}
 Examples of betting functions include the following, also shown in Figure \ref{fig:betfun}:
\vspace{-1mm}
\begin{itemize}
	\item Constant betting functions 
\vspace{-1mm}	
\[
	 \phi^k({\bf x},{\bf c})=\phi^k({\bf x})
\vspace{-1mm}
\]
 for example based on trained classifiers $\phi^k({\bf x},{\bf c})=\eta h^k({\bf x})$, where $\eta\in (0,1]$ is constant.
	\item Linear betting functions 
\vspace{-1mm}	
\begin{equation}\label{eq:linbet}
	\phi^k({\bf x},{\bf c})=(1-c_k)h^k({\bf x})
\vspace{-1mm}
\end{equation}
\item Aggressive betting functions
\vspace{-1mm}	
\begin{equation} \label{eq:aggbet}
	\phi^k({\bf x},{\bf c})=h^k({\bf x})\begin{cases}
	1 &\text{ if } c_k\leq h^k({\bf x})\\
	0 &\text{ if } c_k> h^k({\bf x})+\epsilon\\
	\frac{h^k({\bf x})+\epsilon-c_k}{\epsilon} &\text{ otherwise } 
	\end{cases}
\end{equation}
\item Logistic betting functions: 
\[
\begin{split}
	&\phi_m^1({\bf x},1-c)=(1-c)(x_m^+-\ln(1-c)/B),\\
	&\phi_m^2({\bf x},c)=c(-x_m^--\ln c /B)
	\end{split}
\]
where $x^+=xI(x>0),x^-=xI(x<0)$ and $B=\sum_m \beta_m$.
\end{itemize}

The betting functions play a similar role to the {\em potential functions} from maximum entropy models \citep{berger1996maximum,ratnaparkhi1996maximum,zhu1998frf}, in that they make a conversion from the feature output (or classifier output for some markets) to a common unit of measure (energy for the maximum entropy models and money for the market). 

The contract price does not fluctuate in our setup, instead it is governed by Equation \eqref{eq:budgetcons}. This equation guarantees that at this price, the total amount obtained from selling contracts to the participants is equal to the total amount won by the winning contracts, independent of the outcome.

\vskip -4mm
\begin{figure}[ht]
\centering
\includegraphics[width=8.cm]{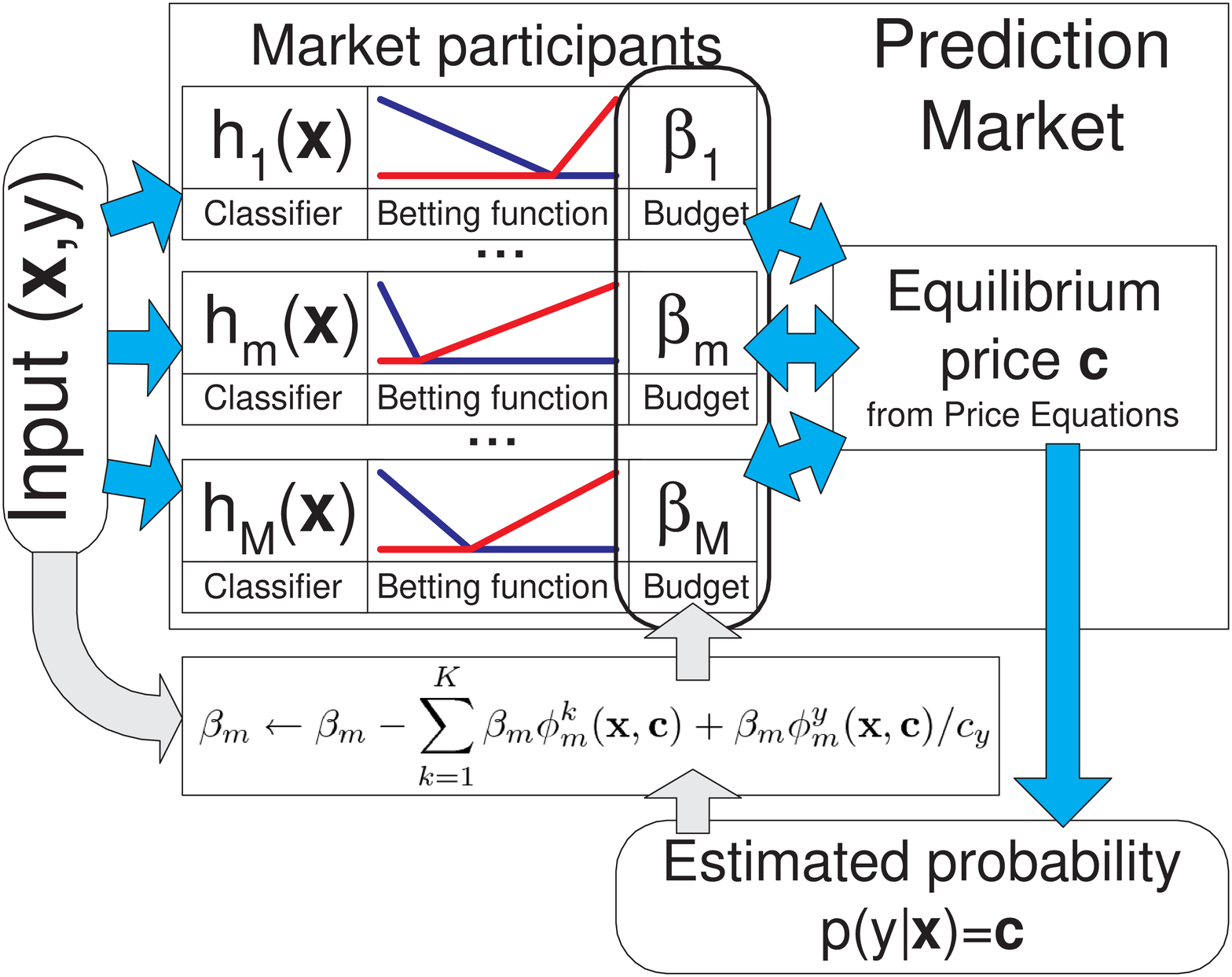}
\vskip -3mm
\caption{Online learning and aggregation using the artificial prediction market. Given feature vector ${\bf x}$, a set of market participants will establish the market equilibrium price $\bf c$, which is an estimator of $P(Y=k|{\bf x})$. The equilibrium price is governed by the Price Equations \eqref{eq:budgetcons}. Online training on an example $({\bf x},y)$ is achieved through {\bf Budget Update} $({\bf x},y,{\bf c})$ shown with gray arrows.}
\label{fig:PredmarketLearn}
\end{figure}
\vskip -2mm
\begin{algorithm}[htb]
   \caption{{\bf Budget Update} $({\bf x},y,{\bf c})$}
   \label{alg:update}
\begin{algorithmic}
   \STATE {\bfseries Input:} Training example $({\bf x},y)$, price {\bf c}
   \FOR{ $m=1$ {\bfseries to} $M$}
   \STATE Update participant $m$'s budget as
\vspace{-2mm}
\begin{equation}
	\beta_m\leftarrow \beta_m-\sum_{k=1}^K \beta_m\phi_m^k({\bf x},{\bf c})+\frac{\beta_m}{c_y}\phi_m^y({\bf x},{\bf c}) \label{eq:budgetupdate}
\vspace{-2mm}
\end{equation}   
   \ENDFOR
\end{algorithmic}
\end{algorithm}
\subsection{Training the Artificial Prediction Market}

Training the market involves initializing all participants with the same budget $\beta_0$ and presenting to the market a set of training examples $({\bf x}_i, y_i), i=1,...,N$. For each example $({\bf x}_i, y_i)$ the participants purchase contracts for the different classes based on the market price $\bf c$ (which is not known yet) and their budgets $\beta_m$ are updated based on the contracts purchased and the true outcome $y_i$. After all training examples have been presented, the participants will have budgets that depend on how well they predicted the correct class $y$ for each training example $\bf x$. This procedure is illustrated in Figure \ref{fig:PredmarketLearn}. 

\begin{algorithm}[htb]
   \caption{{\bf Prediction Market Training}}
   \label{alg:train}
\begin{algorithmic}
   \STATE {\bfseries Input:} Training examples $({\bf x_i},y_i),i=1,...,N$
   \STATE Initialize all budgets $\beta_m=\beta_0,m=1,...,M$.
   \FOR{ each training example $({\bf x}_i,y_i)$}
   \STATE Compute equilibrium price ${\bf c}_i$ using Eq. \ref{eq:budgetcons}
   \STATE Run {\bf Budget Update} $({\bf x}_i,y_i,{\bf c}_i)$
   \ENDFOR
\end{algorithmic}
\end{algorithm}

The budget update procedure subtracts from the budget of each participant the amounts it bets for each class, then rewards each participant based on how many contracts it purchased for the correct class.

Participant $m$ purchased $\beta_m\phi_m^k({\bf x},{\bf c})$ worth of contracts for class $k$, at price $c_k$. Thus the number of contracts purchased for class $k$ is  $\beta_m\phi_m^k({\bf x},{\bf c})/c_k$. Totally, participant $m$'s budget is decreased by the amount $\sum_{k=1}^K \beta_m\phi_m^k({\bf x},{\bf c})$ invested in contracts. Since participant $m$ bought $\beta_m\phi_m^y({\bf x},{\bf c})/c_y$ contracts for the correct class $y$, he is rewarded the amount $\beta_m\phi_m^y({\bf x},{\bf c})/c_y$. 
\vspace{-2mm}


\subsection{The Market Price Equations}

Since we are simulating a real market, we assume that the total amount of money collectively owned by the participants is conserved after each training example is presented. Thus the sum of all participants' budgets $\sum_{m=1}^M \beta_m$ should always be $M\beta_0$, the amount given at the beginning. 
Since any of the outcomes is theoretically possible for each instance, we have the following constraint:
\begin{asum}
The total
budget $\sum_{m=1}^M \beta_m$ must be conserved independent of the outcome $y$. 
\end{asum}

This condition transforms into a set of equations that constrain the market price, which we call the price equations. The market price $\bf c$ also obeys $\sum_{k=1}^K c_k=1$.

Let  $B({\bf x},{\bf c})=\sum_{m=1}^M\sum_{k=1}^K \beta_m\phi_m^k({\bf x},{\bf c})$ be the total bet for observation $\bf x$ at price $\bf c$. We have

\begin{theorem}{\bf Price Equations.} \label{thm:budget} The total budget $\sum_{m=1}^M \beta_m$ is conserved after the {\bf Budget Update}$({\bf x},y,{\bf c})$, independent of the outcome $y$, if and only if $c_k>0, k=1,...,K$ and 
\vspace{-1mm}
\begin{equation}
\sum_{m=1}^M \beta_m\phi_m^k({\bf x},{\bf c})=c_kB({\bf x},{\bf c}), \quad \forall k=1,...,K \label{eq:budgetcons}
\vspace{-1mm}
\end{equation}
\end{theorem}
The proof is given in the Appendix.

\subsection{Price Uniqueness}

The price equations together with the equation $\sum_{k=1}^K c_k=1$ are enough to uniquely determine the market price $\bf c$, under mild assumptions on the betting functions $\phi^k({\bf x},{\bf c})$. 

Observe that if $c_k=0$ for some $k$, then the contract costs $0$ and pays $1$, so there is everything to win. In this case, one should have $\phi^k({\bf x},{\bf c})>0$. 

This suggests a class of betting functions $\phi^k({\bf x},c_k)$ depending only on the price $c_k$ that are continuous and monotonically non-increasing in $c_k$.  If all $\phi_m^k({\bf x},c_k),m=1,...,M$ are continuous and monotonically  non-increasing in $c_k$ with $\phi_m^k({\bf x},0)>0$  then $f_k(c_k)=\frac{1}{c_k}\sum_{m=1}^M \beta_m\phi_m^k({\bf x},c_k) $ is continuous and strictly decreasing in $c_k$ as long as $f_k(c_k)>0$.

To obtain conditions for price uniqueness, we use the following functions
\vspace{-2mm}
\begin{equation}
f_k(c_k)=\frac{1}{c_k}\sum_{m=1}^M \beta_m\phi_m^k({\bf x},c_k), k=1,...,K \label{eq:ckn}
\vspace{-1mm}
\end{equation}
\begin{rem}\label{rem:unique} If all $f_k(c_k)$ are continuous and strictly decreasing in $c_k$ as long as $f_k(c_k)>0$, then for every $n>0$, $n\geq n_k=f_k(1)$ there is a unique $c_k=c_k(n)$ that satisfies $f_k(c_k)=n$.
\end{rem}
The proof is given in the Appendix.

To guarantee price uniqueness, we need at least one market participant to satisfy the following 

\begin{asum}\label{asm:total} The total bet of participant $(\beta_m, \phi_m({\bf x},{\bf c}))$ is positive inside the simplex $\Delta$, i.e. 
\begin{equation} \sum_{j=1}^K \phi_m^j({\bf x},c_j)>0, \; \forall c\in(0,1)^K, \sum_{j=1}^K c_j=1. \label{eq:betfuneq}
\end{equation}
\end{asum}

Then we have the following result, also proved in the Appendix.

\begin{theorem} \label{thm:monbet} Assume all betting functions $\phi_m^k({\bf x},c_k),m=1,...,M, k=1,...,K$ are continuous, with $\phi^k({\bf x},0)>0$ and $\phi_m^k({\bf x},c)/c$ is strictly decreasing in $c$ as long as $\phi_m^k({\bf x},c)>0$. If the betting function $\phi_m({\bf x},{\bf c})$ of least one participant with $\beta_m>0$ satisfies Assumption \ref{asm:total}, then for the {\bf Budget Update}$({\bf x},y,{\bf c})$ there is a unique price ${\bf c}=(c_1,...,c_K)\in (0,1)^K\cap \Delta$ such that the total budget $\sum_{m=1}^M \beta_m$ is conserved.
\end{theorem}

Observe that all four betting functions defined in Section \ref{sec:setup} ( constant, linear, aggressive and logistic) satisfy the conditions of Theorem \ref{thm:monbet}, so there is a unique price that conserves the budget.

\subsection{Solving the Market Price Equations}

In practice, a double bisection algorithm could be used to find the equilibrium price, computing each $c_k(n)$ by the bisection method, and employing another bisection algorithm to find $n$ such that the price condition $\sum_{k=1}^K c_k(n)=1$ holds. Observe that the $n$ satisfying $\sum_{k=1}^K c_k(n)=1$ can be bounded from above by
\vspace{-1mm}
\[
	n=n\sum_{k=1}^K c_k(n)=\sum_{k=1}^K c_k(n)f_k(c_k(n))=\sum_{k=1}^K \sum_{m=1}^M\beta_m\phi_m^k({\bf x},{\bf c})\leq \sum_{m=1}^M \beta_m
\vspace{-1mm}
\]
because for each $m$, $\sum_{k=1}^K \phi_m^k({\bf x},{\bf c})\leq 1$.

A potentially faster alternative to the double bisection method is the Mann Iteration \citep{mann1953} described in Algorithm \ref{alg:mannit}. The price equations can be viewed as fixed point equation $F({\bf c})={\bf c}$, where $F({\bf c})=\frac{1}{n}(f_1({\bf c}),...,f_K({\bf c}))$ with $f_k({\bf c})=\sum_{m=1}^m \beta_m \phi_m^k({\bf x},c_k)$.
The Mann iteration is a fixed point algorithm, which makes weighted update steps 
\[
{\bf c}^{t+1}=(1-\frac{1}{t}){\bf c}^{t}+\frac{1}{t} F({\bf c}^{t})
\]

The Mann iteration is guaranteed to converge for contractions or pseudo-contractions. However, we observed experimentally that it usually converges in only a few (up to 10) steps, making it about 100-1000 times faster than the double bisection algorithm. If, after a small number of steps, the Mann iteration has not converged, the double bisection algorithm is used on that instance to compute the equilibrium price. However, this happens on less than $0.1\%$ of the instances.

\begin{algorithm}[ht]
  \caption{{\bf Market Price by Mann Iteration}}
  \label{alg:mannit}
\begin{algorithmic}
  \STATE Initialize $i = 1,\ c_k = \frac{1}{K},k=1,...,K$
  \REPEAT
    \STATE $f_k = \sum_m \beta_m \phi_m^k(x,\mathbf{c})$
    \STATE $n = \sum_k f_k$
    \IF{$n \neq 0$}
      \STATE $f_k \leftarrow \frac{f_k}{n}$
      \STATE $r_k = f_k - c_k$
      \STATE $c_k \leftarrow \frac{(i-1) c_k + f_k}{i}$
    \ENDIF
    \STATE $i \leftarrow i+1$
  \UNTIL{$\sum_k |r_k| \leq \epsilon$ or $n = 0$ or $i > i_{\text{max}}$}
\end{algorithmic}
\end{algorithm}

\subsection{Two-class Formulation}
\vskip 3mm
For the two-class problem, i.e. $K=2$, the budget equation can be simplified by writing ${\bf c}=(1-c,c)$ and obtaining the {\em two-class market price equation}
\vspace{-2mm}
\begin{equation}
(1-c)\sum_{m=1}^M \hspace{-1mm}\beta_m\phi_m^2({\bf x},c)-c\sum_{m=1}^M \hspace{-1mm}\beta_m\phi_m^1({\bf x},1-c)=0 \label{eq:2class}
\vspace{-2mm}
\end{equation}
This can be solved numerically directly in $c$ using the bisection method. Again, the solution is unique if $\phi_m^k({\bf x},c_k),m=1,...,M, k=1,2$ are continuous, monotonically  non-increasing and obey condition \eqref{eq:betfuneq}. Moreover, the solution is guaranteed to exist if there exist $m,m'$ with $\beta_m>0, \beta_{m'}>0$ and such that $\phi_m^2({\bf x},0)>0, \phi_{m'}^1({\bf x},1)>0$.

\section{Relation to Existing Supervised Learning Methods}
\vspace{-1mm}
There is a large degree of flexibility in choosing the betting functions $\phi_m({\bf x},{\bf c})$. Different betting functions give different ways to fuse the market participants. In what follows we prove that by choosing specific betting functions, the artificial prediction market behaves like a linear aggregator or logistic regressor, or that it can be used as a kernel-based classifier.

\subsection{Constant Betting and Linear Aggregation}\label{sec:ctbet}

For markets with constant betting functions, $\phi_m^k({\bf x},{\bf c})=\phi_m^k({\bf x})$ the market price has a simple analytic formula, proved in the Appendix.

\begin{theorem}{\bf Constant Betting.}\label{thm:ctbet}
If all betting function are constant $\phi_m^k({\bf x},{\bf c})= \phi_m^k({\bf x})$, then the equilibrium price is
\vspace{-1mm}
\begin{equation}
{\bf c}=\frac{\sum_{m=1}^M \beta_m\phi_m({\bf x})}{\sum_{m=1}^M\sum_{k=1}^K \beta_m\phi_m^k({\bf x})} \label{eq:ctbet}
\vspace{-1mm}
\end{equation}
Furthermore, if the betting functions are based on classifiers $\phi_m^k({\bf x},{\bf c})= \eta h_m^k({\bf x})$ then the  equilibrium price is obtained by linear aggregation
\vspace{-1mm}
\begin{equation}
{\bf c}=\frac{\sum_{m=1}^M \beta_mh_m({\bf x})}{\sum_{m=1}^M \beta_m}=\sum_m \alpha_m h_m({\bf x}) \label{eq:ctclf}
\vspace{-1mm}
\end{equation}
\end{theorem}

This way the artificial prediction market can model linear aggregation of classifiers. Methods such as Adaboost \citep{freund1996experiments,friedman2000alr,schapire2003bam} and Random Forest \citep{breiman_random_2001} also aggregate their constituents using linear aggregation. However, there is more to Adaboost and Random Forest than linear aggregation, since it is very important how to construct the constituents that are aggregated. 

In particular, the random forest \citep{breiman_random_2001} can be viewed as an artificial prediction market with constant betting (linear aggregation) where all participants are random trees with the same budget $\beta_m=1,m=1,...,M$.

We also obtain an  analytic form of the budget update:
\vspace{-1mm}
\[
\beta_m\leftarrow \beta_m -\beta_m\sum_{k=1}^K \phi_m^k({\bf x})+ \beta_m\frac{\phi_m^y({\bf x})\sum_{j=1}^M \sum_{k=1}^K \beta_j\phi_j^k({\bf x})}{\sum_{j=1}^M \beta_j \phi_j^y({\bf x})}
\vspace{-1mm}
\]
which for classifier based betting functions $\phi_m^k({\bf x},{\bf c})= \eta h_m^k({\bf x})$ becomes:
\vspace{-1mm}
\[
\beta_m\leftarrow \beta_m (1-\eta)+\eta \beta_m\frac{h_m^y({\bf x})\sum_{j=1}^M \beta_j }{\sum_{j=1}^M \beta_j h_j^y({\bf x})} 
\vspace{-1mm}
\]
This is a novel online update rule for linear aggregation.

\subsection{Prediction Markets for Logistic Regression}\label{sec:logbet}

A variant of logistic regression can also be modeled using prediction markets, with the following betting functions
\[
\begin{split}
	&\phi_m^1({\bf x},1-c)=(1-c)(x_m^+-\frac{1}{B}\ln(1-c)),\\
	&\phi_m^2({\bf x},c)=c(-x_m^--\frac{1}{B}\ln c )
\end{split}
\]
where $x^+=xI(x>0),x^-=xI(x<0)$ and $B=\sum_m \beta_m$. The two class equation \eqref{eq:2class} becomes:
$	\sum_{m=1}^M \beta_m c(1-c)(x_m-\ln(1-c)/B+\ln c/B)=0$ so
$\ln \frac{1-c}{c} = \sum_{m=1}^M \beta_m x_m$,
which gives the logistic regression model
\vspace{-2mm}
\[
\hat p(Y=1|{\bf x})=c=\frac{1}{1+\exp( \sum_{m=1}^M \beta_m {\bf x}_m)}
\vspace{-1mm}
\]

The budget update equation $\beta_m\leftarrow \beta_m - \eta\beta_m\left[(1-c)x_m^++cx_m^--H(c)/B\right ]+\eta\beta_m u_y(c)$ is obtained,
where $u_1(c)=x_m^+-\ln(1-c)/B,u_2(c)=-x_m^--\ln(c)/B$. 

Writing ${\bf x}\beta= \sum_{m=1}^M \beta_m {\bf x}_m$, the budget update can be rearranged to
\begin{equation}
\beta_m\leftarrow \beta_m-\eta\beta_m \left (x_m-\frac{{\bf x}\beta}{B}\right )\left(y-\frac{1}{1+\exp({\bf x}\beta)}\right).\label{eq:logregmkt}
\end{equation}

This equation resembles the standard per-observation update equation for online logistic regression:
\begin{equation}
\beta_m\leftarrow \beta_m-\eta x_m\left(y-\frac{1}{1+\exp({\bf x}\beta)}\right),\label{eq:logreg}
\vspace{-1mm}
\end{equation}

with two differences. The term ${\bf x}\beta/B$ ensures the budgets always sum to $B$ while the factor $\beta_m$ makes sure that $\beta_m\geq 0$.

The update from eq. \eqref{eq:logregmkt}, like eq.  \eqref{eq:logreg} tries to increase $|{\bf x}\beta|$, but it does that subject to constraints that $\beta_m\geq 0, m=1,...,M$ and  $\sum_{m=1}^M \beta_m=B$. Observe also that multiplying $\beta$ by a constant does not change the decision line of the logistic regression.

\subsection{Relation to Kernel Methods}\label{sec:svm}

Here we construct a market participant from each training example $(x_n,y_n), n=1,...N$, thus the number of participants $M$ is the number $N$ of training examples. We construct a participant from training example $(x_m,y_m)$ by defining the following betting functions in terms of $u_m({\bf x})=\frac{{\bf x}_m^T{\bf x}}{\|{\bf x}_m\|\|{\bf x}\|}$:
\vspace{-1mm}
\begin{equation}
\begin{split}
\label{eq:svmbetfun}
	&\phi_m^{y_m}({\bf x})=u_m({\bf x})^+=\begin{cases} u_m({\bf x}) \text { if } u_m({\bf x})\geq 0\\
	0 \text{ else}
	\end{cases},\\
	&\phi_m^{2-y_m}({\bf x})=-u_m({\bf x})^-=\begin{cases}
	0 \text { if } u_m({\bf x})\geq 0\\
	-u_m({\bf x})  \text{ else}
	\end{cases}
\end{split}
\end{equation}
Observe that these betting functions do not depend on the contract price $c$, so it is a constant market but not one based on classifiers. The two-class price equation gives
\vspace{-0mm}
\[
	c=\frac{\displaystyle \sum_m \beta_m \phi_m^2({\bf x})}{\displaystyle\sum_m \beta_m (\phi_m^1({\bf x})+\phi_m^2({\bf x}))}\hspace{-0.5mm}=\hspace{-0.5mm}
	\frac{\displaystyle\sum_m \hspace{-0.5mm}\beta_m [y_mu_m\hspace{-0.5mm}({\bf x})\hspace{-0.5mm}-\hspace{-0.5mm}u_m({\bf x})^-]}{\displaystyle\sum_m \beta_m |u_m({\bf x})|}
\vspace{-0mm}
\]
since it can be verified that $\phi_m^2({\bf x})=y_mu_m({\bf x})-u_m({\bf x})^-$ and $\phi_m^1({\bf x})+\phi_m^2({\bf x})=|u_m({\bf x})|$.

The decision rule $c>0.5$ becomes $\sum_m \beta_m \phi_m^2({\bf x})>\sum_m \beta_m \phi_m^1({\bf x})$ or $\sum_m \beta_m (\phi_m^2({\bf x})-\phi_m^1({\bf x}))>0$. Since $\phi_m^2({\bf x})-\phi_m^1({\bf x})=(2y_m-2)u_m({\bf x})=(2y_m-2)\frac{{\bf x}_m^T{\bf x}}{\|{\bf x}_m\|\|{\bf x}\|}$ (since in our setup $y_m\in \{1,2\}$), we obtain the SVM type of decision rule with $\alpha_m=\beta_m/\|{\bf x}_m\|$: 
\[
	h({\bf x})=\text{sgn}(\sum_{m=1}^M \alpha_m (2y_m-3) {\bf x}_m^T {\bf x})
\]
The budget update becomes in this case:
\[
\beta_m\leftarrow \beta_m-\eta\beta_m |u_m({\bf x})|+\eta\beta_m \frac{\phi_m^y({\bf x})}{c_y}
\]

\begin{figure*}[ht]
\centering
\vspace{-1mm}
\includegraphics[height=5.cm]{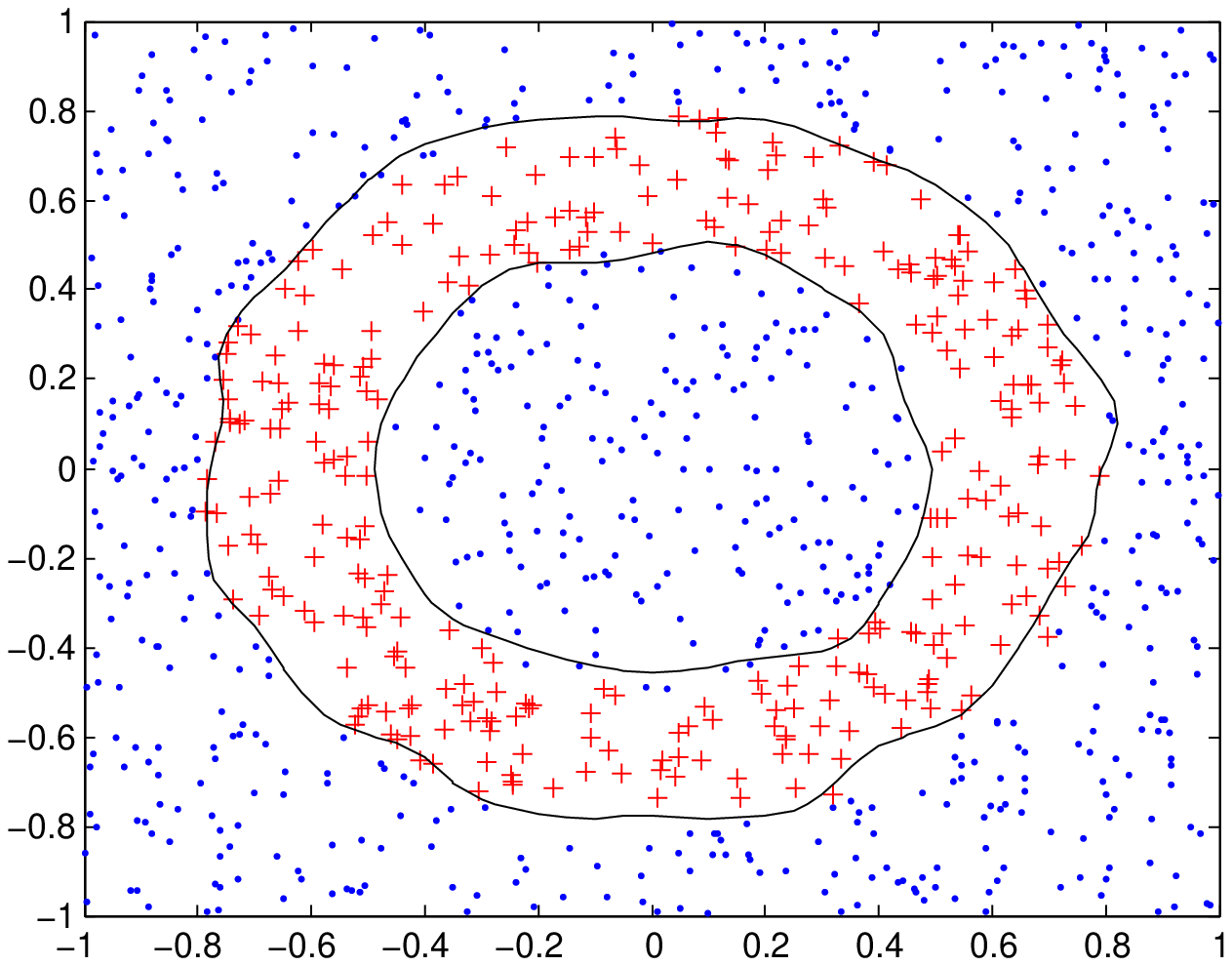}
\includegraphics[height=5cm]{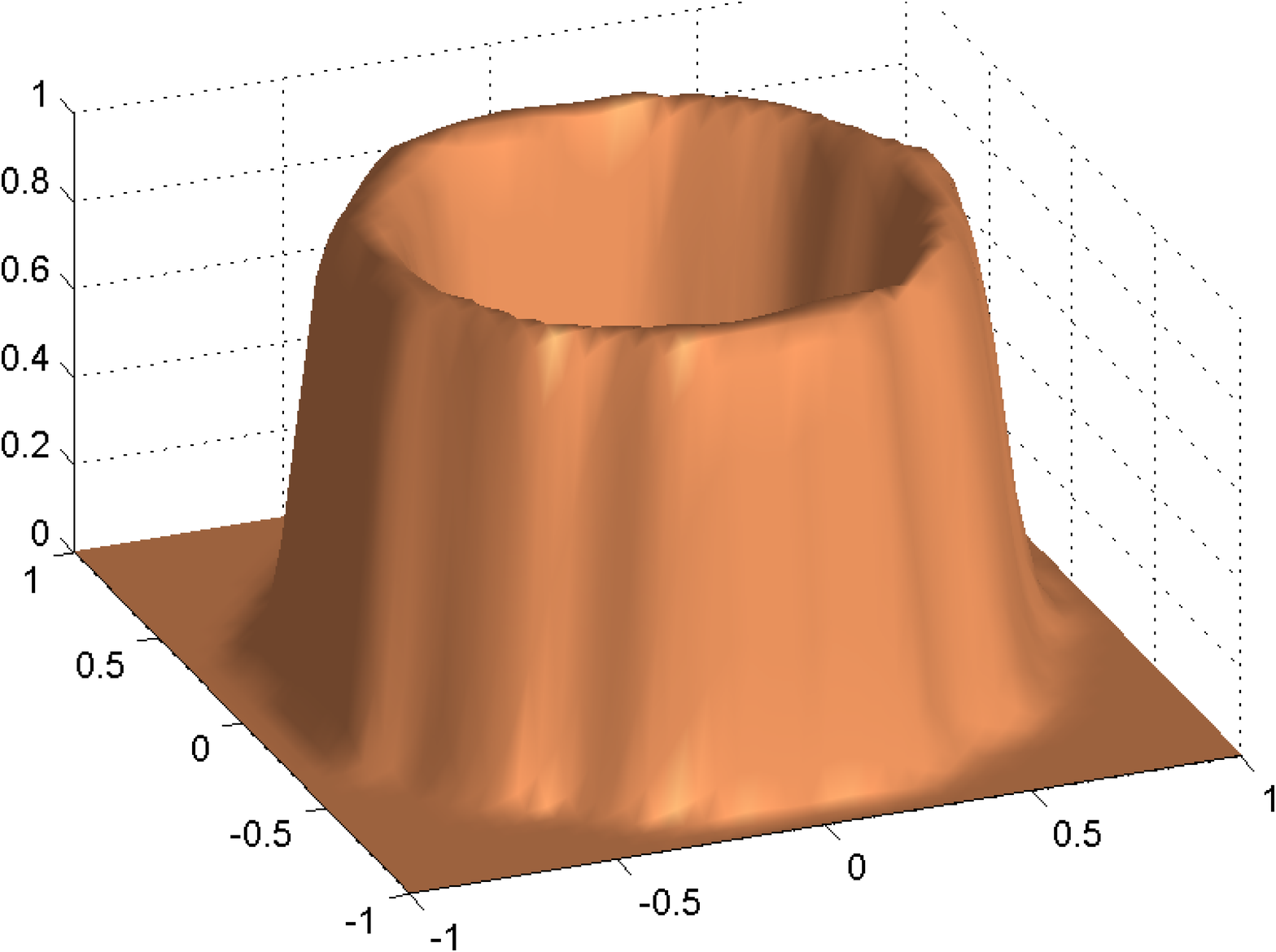}
\vskip -3mm
\caption{Left: 1000 training examples and learned decision boundary (right) for an RBF kernel-based market from eq. \eqref{eq:svmbetfun} with $\sigma=0.1$. Right: estimated probability function.}
\label{fig:svmdemo}
\end{figure*}

The same reasoning carries out for $u_m({\bf x})=K({\bf x}_m,{\bf x})$ with the RBF kernel $K({\bf x}_m,{\bf x})=\exp(-\|{\bf x}_m-{\bf x}\|^2/\sigma^2)$.
In Figure \ref{fig:svmdemo}, left, is shown an example of the decision boundary of a market trained online with an RBF kernel with $\sigma=0.2$ on 1000 examples uniformly sampled in the $[-1,1]^2$ interval. In Figure \ref{fig:svmdemo}, right is shown the estimated probability $\hat p(y=1|{\bf x})$.

This example shows that the artificial prediction market is an online method with enough modeling power to represent complex decision boundaries such as those given by RBF kernels through the betting functions of the participants. It will be shown in Theorem \ref{thm:ctloss} that the constant market maximizes the likelihood, so it is not clear yet what can be done to obtain a small number of support vectors as in the online kernel-based methods \citep{bordes2005fast,cauwenberghs2001incremental,kivinen2004online}. 

\section{Prediction Markets and Maximum Likelihood}

This section discusses what type of optimization is performed during the budget update from eq. \eqref{eq:budgetupdate}. Specifically, we prove that the artificial prediction markets perform maximum likelihood learning of the parameters by a version of gradient ascent.

Consider the reparametrization $\gamma=(\gamma_1,...,\gamma_M)=(\sqrt{\beta_1},...,\sqrt{\beta_M})$.
The  market price ${\bf c}({\bf x})=(c_1({\bf x}),...,c_K({\bf x})$ is an estimate of the class probability $p(y=k|{\bf x})$ for each instance ${\bf x}\in \Omega$. Thus a set of training observations $({\bf x}_i,y_i),i=1,...,N$, since $\hat p(y=y_i|{\bf x}_i)=c_{y_i}({\bf x}_i)$, the (normalized) log-likelihood function is
\vspace{-2mm}
\begin{equation}
L(\gamma)=\frac{1}{N}\sum_{i=1}^N \ln \hat p(y=y_i|{\bf x}_i)=\frac{1}{N}\sum_{i=1}^N \ln c_{y_i}({\bf x}_i)\label{eq:lik}
\vspace{-2mm}
\end{equation}

We will again use the total amount bet $B({\bf x},{\bf c})=\sum_{m=1}^M\sum_{k=1}^K \beta_m\phi_m^k({\bf x},{\bf c})$  for observation $\bf x$ at  market price $\bf c$. 

We will first focus on the constant market $\phi_m^k({\bf x},{\bf c})=\phi_m^k({\bf x})$, in which case  $B({\bf x},{\bf c})=B({\bf x})=\sum_{m=1}^M\sum_{k=1}^K \beta_m\phi_m^k({\bf x})$. We introduce a batch update on all the training examples $({\bf x}_i,y_i),i=1,...,N$:
\vspace{-1mm}
\begin{equation}
	\beta_m\leftarrow \beta_m +\beta_m \frac{\eta}{N}\sum_{i=1}^N\frac{1}{B({\bf x}_i)}
\left ( \frac{\phi^{y_i}_m({\bf x}_i)}{c_{y_i}({\bf x}_i)}  -\sum_{k=1}^K\phi^k_m({\bf x}_i)\right ). \label{eq:ctbatchupd}
\vspace{-1mm}
\end{equation}
Equation \eqref{eq:ctbatchupd} can be viewed as presenting all observations $({\bf x}_i,y_i)$ to the market simultaneously instead of sequentially.
The following statement is proved in the Appendix
\begin{theorem}{\bf ML for constant market.}\label{thm:ctloss}
The update \eqref{eq:ctbatchupd} for the constant market maximizes the likelihood \eqref{eq:lik} by gradient ascent on $\gamma$ subject to the constraint $\sum_{m=1}^M \gamma_m^2=1$. The incremental update 
\vspace{-1mm}
\begin{equation}
	\beta_m\leftarrow \beta_m +\beta_m\frac{\eta}{B({\bf x}_i)}
\left ( \frac{\phi^{y_i}_m({\bf x}_i)}{c_{y_i}({\bf x}_i)}  -\sum_{k=1}^K\phi^k_m({\bf x}_i)\right ). \label{eq:ctincupd}
\vspace{-1mm}
\end{equation}

maximizes the likelihood \eqref{eq:lik} by constrained stochastic gradient ascent.
\end{theorem}

In the general case of non-constant betting functions, the  log-likelihood is
\begin{equation}
L(\gamma)=\sum_{i=1}^N\log c_{y_i}({\bf x}_i)=\sum_{i=1}^N\log \sum_{m=1}^M \gamma_m^2 \phi^{y_i}_m({\bf x}_i,{\bf c}({\bf x}_i))-\sum_{i=1}^N\log \sum_{k=1}^K \sum_{m=1}^M \gamma_m^2 \phi^k_m({\bf x}_i,{\bf c}({\bf x}_i))\label{eq:logc}
\end{equation}
If we ignore the dependence of $\phi^k_m({\bf x}_i,{\bf c}({\bf x}_i))$ on $\gamma$ in \eqref{eq:logc}, and approximate the
gradient as:
\vspace{-2mm}
\[
\frac{\partial L(\gamma)}{\partial \gamma_j}\approx \sum_{i=1}^N\left(\frac{\gamma_j\phi^{y_i}_j({\bf x}_i,{\bf c}({\bf x}_i))}{\sum_{m=1}^M \gamma_m^2 \phi^{y_i}_m({\bf x}_i,{\bf c}({\bf x}_i))}-\frac{\gamma_j\sum_{k=1}^K\phi^k_j({\bf x}_i,{\bf c}({\bf x}_i))}{\sum_{k=1}^K\sum_{m=1}^M \gamma_m^2 \phi^k_m({\bf x}_i,{\bf c}({\bf x}_i))}\right)
\vspace{-1mm}
\]
\noindent then the proof of Theorem \ref{thm:ctloss} follows through and we obtain the following market update
\begin{equation}
	\beta_m\leftarrow \beta_m+\beta_m\frac{\eta}{B({\bf x},{\bf c})}\left[ \frac{\phi_m^y({\bf x},{\bf c})}{c_y}-\sum_{k=1}^K \phi_m^k({\bf x},{\bf c})\right ], \; m=1,...,M \label{eq:budgetupd1}
\end{equation}

This way we obtain only an approximate statement in the general case
\begin{rem}{\bf Maximum Likelihood.}\label{thm:lossfun}
The prediction market update \eqref{eq:budgetupd1} finds an approximate maximum of the likelihood \eqref{eq:lik} subject to the constraint $\sum_{m=1}^M \gamma_m^2=1$ by an approximate constrained stochastic gradient ascent. 
\end{rem}

Observe that the updates from \eqref{eq:ctincupd} and \eqref{eq:budgetupd1} differ from the update \eqref{eq:budgetupdate} by using an adaptive step size $\eta/B({\bf x},{\bf c})$ instead of the fixed step size $1$.

It is easy to check that maximizing the likelihood is equivalent to minimizing an approximation of the expected KL divergence to the true distribution 
\[
E_\Omega[KL\left( p(y|{\bf x}),c_y({\bf x})\right) ]=\int_\Omega p({\bf x}) \int_Y p(y|{\bf x}) \log \frac {p(y|{\bf x})}{c_y({\bf x})} dy d{\bf x}
\]
obtained using the training set as Monte Carlo samples from $p({\bf x},y)$.

In many cases the number of negative examples is much larger than the positive examples, and is desired to maximize a weighted log-likelihood
\[
L(\gamma)=\frac{1}{N}\sum_{i=1}^N w({\bf x}_i)\ln c_{y_i}({\bf x}_i)
\vspace{-2mm}
\]
This can be achieved (exactly for constant betting and approximately in general) using the weighted update rule
\begin{equation}
	\beta_m\leftarrow \beta_m+\eta w({\bf x})\frac{\beta_m}{B({\bf x},{\bf c})}\left[ \frac{\phi_m^y({\bf x},{\bf c})}{c_y}-\sum_{k=1}^K \phi_m^k({\bf x},{\bf c})\right ], \; m=1,...,M \label{eq:wtupdate}
\end{equation}

The parameter $\eta$ and the number of training epochs can be used to control how close the budgets $\beta$ are to the ML optimum, and this way avoid overfitting the training data.
 

An important issue for the real prediction markets is the {\em efficient market hypothesis}, which states that the market price fuses in an optimal way the information available to the market participants \citep{fama1970efficient,basu1977investment,malkiel2003efficient}.
From Theorem \ref{thm:ctloss} we can draw the following conclusions for the artificial prediction market with constant betting:
\begin{enumerate}
\item In general, an untrained market (in which the budgets have not been updated based on training data) will not satisfy the efficient market hypothesis.
\item The market trained with a large amount of representative training data and small $\eta$ satisfies the efficient market hypothesis.
\end{enumerate}

\section{Specialized Classifiers}

The prediction market is capable of fusing the information available to the market participants, which can be trained classifiers. 
These classifiers are usually suboptimal, due to computational or complexity constraints, to the way they are trained, or other reasons.  

In boosting, all selected classifiers are aggregated for each instance $\bf x\in \Omega$. This can be detrimental since some classifiers could perform poorly on subregions of the instance space $\Omega$, degrading the performance of the boosted classifier. In many situations there exist simple rules that hold on subsets of $\Omega$ but not on the entire $\Omega$. Classifiers trained on such subsets $D_i\subset \Omega$, would have small misclassification error on $D_i$ but unpredictable behavior outside of $D_i$. The artificial prediction market can aggregate such classifiers, transformed into participants that don't bet anything outside of their domain of expertise $D_i\subset \Omega$. This way, for different instances $\bf x\in \Omega$, different subsets of participants will contribute to the resulting probability estimate. We call these {\em specialized classifiers} since they only give their opinion through betting on observations that fall inside their domain of specialization.

Thus a specialized classifier with a domain $D$ would have a betting function of the form:
\begin{equation}
\phi^k({\bf x},{\bf c})=\begin{cases} \varphi^k({\bf x},{\bf c}) \text{ if }{\bf x}\in D\\ \label{eq:spec}
0 \text{ else}
\end{cases}
\end{equation}

\begin{figure}[ht]
\centering
\includegraphics[width=5.5cm]{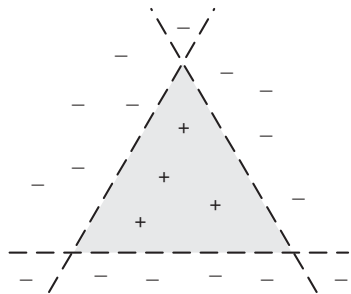}
\vskip -4mm
\caption{A perfect classifier can be constructed for the triangular region above from a market of six specialized classifiers that only bid on a half-plane determined by one side of the triangle. Three of these specialized classifiers have 100\% accuracy while the other three have low accuracy. Nevertheless, the market is capable of obtaining 100\% overall accuracy.}
\label{fig:triangle}
\end{figure}

This idea is illustrated on the following simple 2D example of a triangular region, shown in Figure \ref{fig:triangle}, with positive examples inside the triangle and negatives outside. An accurate classifier for that region can be constructed using six market participants, one for each half-plane determined by each side of the triangle.

Three of these classifiers correspond to the three half planes that are outside the triangle. These participants have 100\% accuracy in predicting the observations, all negatives, that fall in their half planes and don't bet anything outside of their half planes. The other three classifiers are not very good, and will have smaller budgets. On an observation that lies outside of the triangle, one or two of the high-budget classifiers will bet a large amount on the correct prediction and will drive the output probability. When an observation falls inside the triangle, only the small-budget classifiers will participate but will be in agreement and still output the correct probability. Evaluating this market on 1000 positives and 1000 negatives showed that the market obtained a prediction accuracy of 100\%.

There are many ways to construct specialized classifiers, depending on the problem setup. In natural language processing for example, a specialized classifier could be based on grammar rules, which work very well in many cases, but not always. 

We propose two generic sets of specialized classifiers. The first set are the leaves of the random trees of a random forest while the second set are the leaves of the decision trees trained by adaboost. Each leaf $f$ is a rule that defines a domain $D_f=\{ {\bf x} \in \Omega, f({\bf x})=1\}$ of the instances that obey that rule. The betting function of this specialized classifier is given in eq. \eqref{eq:spec} where $\varphi_f^k({\bf x},{\bf c})$ is based on the associated classifier $h^k_f({\bf x})=n_{fk}/n_f$, obtaining constant, linear and aggressive versions. Here  $n_{fk}$ is the number of training instances of class $k$ that obey rule $f$ and $n_f=\sum_k n_{fk}$. By the way the random trees are trained, usually $n_f=n_{fk} $ for some k. 

In \cite{friedman2008statistics} these rules were combined using a linear aggregation method similar to boosting. One could also use other nodes of the random tree, not necessarily the leaves, for the same purpose. 

It can be verified using eq. \eqref{eq:ctbet} that constant specialized betting is the linear aggregation of the participants that are currently betting. This is different than the linear aggregation of all the classifiers.

\section{Related Work}

This work borrows prediction market ideas from Economics and brings them to Machine Learning for supervised aggregation of classifiers or features in general.

{\bf Related work in Economics.} Recent work in Economics \citep{manski2006interpreting,perols2009information,plott2003parimutuel} investigates the information fusion of the prediction markets. However, none of these works aims at using the prediction markets as a tool for learning class probability estimators in a supervised manner.

Some works \citep{perols2009information,plott2003parimutuel} focus on parimutuel betting mechanisms for combining classifiers. In parimutuel betting contracts are sold for all possible outcomes (classes) and the entire budget (minus fees) is divided between the participants that purchased contracts for the winning outcome. Parimutuel betting has a different way of fusing information than the Iowa prediction market.

The information based decision fusion \citep{perols2009information} is a first version of an artificial prediction market. It aggregates classifiers through the parimutuel betting mechanism, using a loop that updates the odds for each outcome and takes updated bets until convergence. This insures a stronger information fusion than without updating the odds. Our work is different in many ways. First our work uses the Iowa electronic market instead of parimutuel betting with odds-updating. Using the Iowa model allowed us to obtain a closed form equation for the market price in some important cases. It also allowed us to relate the market to some existing learning methods. Second, our work presents a multi-class formulation of the prediction markets as opposed to a two-class approach presented in \citep{perols2009information}. 
Third, the analytical market price formulation allowed us to prove that the constant market performs maximum likelihood learning.
Finally, our work evaluates the prediction market not only in terms of classification accuracy but also in the accuracy of predicting the exact class conditional probability given the evidence.


{\bf Related work in Machine Learning.} 
Implicit online learning \citep{kulis2010implicit} presents a generic online learning method that balances between a ``conservativeness'' term that discourages large changes in the model and a ``correctness'' term that tries to adapt to the new observation. Instead of using a linear approximation as other online methods do, this approach solves an implicit equation for finding the new model. In this regard, the prediction market also solves an implicit equation at each step for finding the new model, but does not balance two criteria like the implicit online learning method. Instead it performs maximum likelihood estimation, which is consistent and asymptotically optimal. In experiments, we observed that the prediction market obtains significantly smaller misclassification errors on many datasets compared to implicit online learning.

Specialization can be viewed as a type of reject rule \citep{chow1970optimum,tortorella2004reducing}. However, instead of having a reject rule for the aggregated classifier, each market participant has his own reject rule to decide on what observations to contribute to the aggregation. ROC-based reject rules \citep{tortorella2004reducing} could be found for each market participant and used for defining its domain of specialization.
Moreover, the market can give an overall reject rule on hopeless instances that fall outside the specialization domain of all participants. No participant will bet for such an instance and this can be detected as an overall rejection of that instance.

If the overall reject option is not desired, one could avoid having instances for which no classifiers bet by including in the market a set of participants that are all the leaves of a number of random trees. This way, by the design of the random trees, it is guaranteed that each instance will fall into at least one leaf, i.e. participant, hence the instance will not be rejected.

A simplified specialization approach is taken in delegated classifiers \citep{ferri2004delegating}. A first classifier would decide on the relatively easy instances and would delegate more difficult examples to a second classifier. This approach can be seen as a market with two participants that are not overlapping. The specialization domain of the second participant is defined by the first participant. The market takes a more generic approach where each classifier decides independently on which instances to bet. 

The same type of leaves of random trees (i.e. rules) were used by \cite{friedman2008statistics} for linear aggregation. However, our work presents a more generic aggregation method through the prediction market, with linear aggregation as a particular case, and we view the rules as one sort of specialized classifiers that only bid in a subdomain of the feature space.  


Our earlier work \citep{lay2010apm} focused only on aggregation of classifiers and did not discuss the connection between the artificial prediction markets and logistic regression, kernel methods and maximum likelihood learning. Moreover, it did not include an experimental comparison with implicit online learning and adaboost. 

Two other prediction market mechanisms have been recently proposed in the literature. The first one \citep{chen2010new,chen2011optimization} has the  participants entering the market sequentially. Each participant is paid by an entity called the market maker according to a predefined scoring rule. The second prediction market mechanism is the machine learning market \citep{amos2011machine, storkey2012isoelastic}, dealing with all participants simultaneously. Each market participant purchases contracts for the possible outcomes to maximize its own utility function. The equilibrium price of the contracts is computed by an optimization procedure. Different utility functions result in different forms of the equilibrium price, such as the mean, median, or geometric mean of the participants' beliefs.

\section{Experimental Validation}

In this section we present experimental comparisons of the performance of different artificial prediction markets with random forest, adaboost and implicit online learning \citep{kulis2010implicit}. 

Four artificial prediction markets are evaluated in this section. These markets have the same classifiers, namely the leaves of the trained random trees, but differ either in the betting functions or in the way the budgets are trained as follows:
\begin{enumerate}
	\item The first market has constant betting and equal budgets for all participants. We proved in Section \ref{sec:ctbet} that this is a random forest \citep{breiman_random_2001}.
	\item The second market has constant betting based on specialized classifiers (the leaves of the random trees), with the budgets initialized with the same values like the market 1 above, but trained using the update equation \eqref{eq:ctincupd}. Thus after training it will be different from market 1.
	\item The third market has linear betting functions \eqref{eq:linbet}, for which the market price can be computed analytically only for binary classification. The market is initialized with equal budgets and trained using eq. \eqref{eq:budgetupd1}.
	\item The fourth market has aggressive betting \eqref{eq:aggbet} with $\epsilon=0.01$ and the market price computed using the Mann iteration Algorithm \ref{alg:mannit}. The market is initialized with equal budgets and trained using eq. \eqref{eq:budgetupd1}. The value $\epsilon=0.01$ was chosen for simplicity; a better choice would be to obtain it by cross-validation. 
\end{enumerate}

For each dataset, 50 random trees are trained on bootstrap samples of the training data. These trained random trees are used to construct the random forest and the other three markets described above. This way only the aggregation capabilities of the different markets are compared.

The budgets in the markets 2-4 described above are trained on the same training data using the update equation \eqref{eq:budgetupd1} which simplifies to \eqref{eq:ctincupd} for the constant market.

A C++ implementation of these markets can be found at the following address:
\vspace{-5mm}
\begin{verbatim}
http://stat.fsu.edu/~abarbu/Research/PredMarket.zip
\end{verbatim}

\subsection{Case Study}

We first investigate the behavior of three markets on a dataset in terms of training and test error as well as loss function. For that, we chose the \verb satimage  dataset from the UCI repository \citep{blake1998uci} since it has a supplied test set. The  \verb satimage  dataset has a training set of size $4435$ and a test set of size $2000$.

The markets investigated are the constant market with both incremental and batch updates, given in eq. \eqref{eq:ctincupd} and \eqref{eq:ctbatchupd} respectively, the linear and aggressive  markets with incremental updates given in \eqref{eq:budgetupd1}. Observe that the $\eta$ in eq. \eqref{eq:ctincupd} is not divided by $N$ (the number of observations) while the $\eta$ in  \eqref{eq:ctbatchupd} is divided by $N$. Thus to obtain the same behavior the $\eta$ in \eqref{eq:ctincupd} should be the $\eta$ from  \eqref{eq:ctbatchupd} divided by $N$. We used $\eta=100/N$ for the incremental update and $\eta=100$ for the batch update unless otherwise specified.

\begin{figure}[ht]
\centering
\hspace{-4mm}\includegraphics[height=3.95cm]{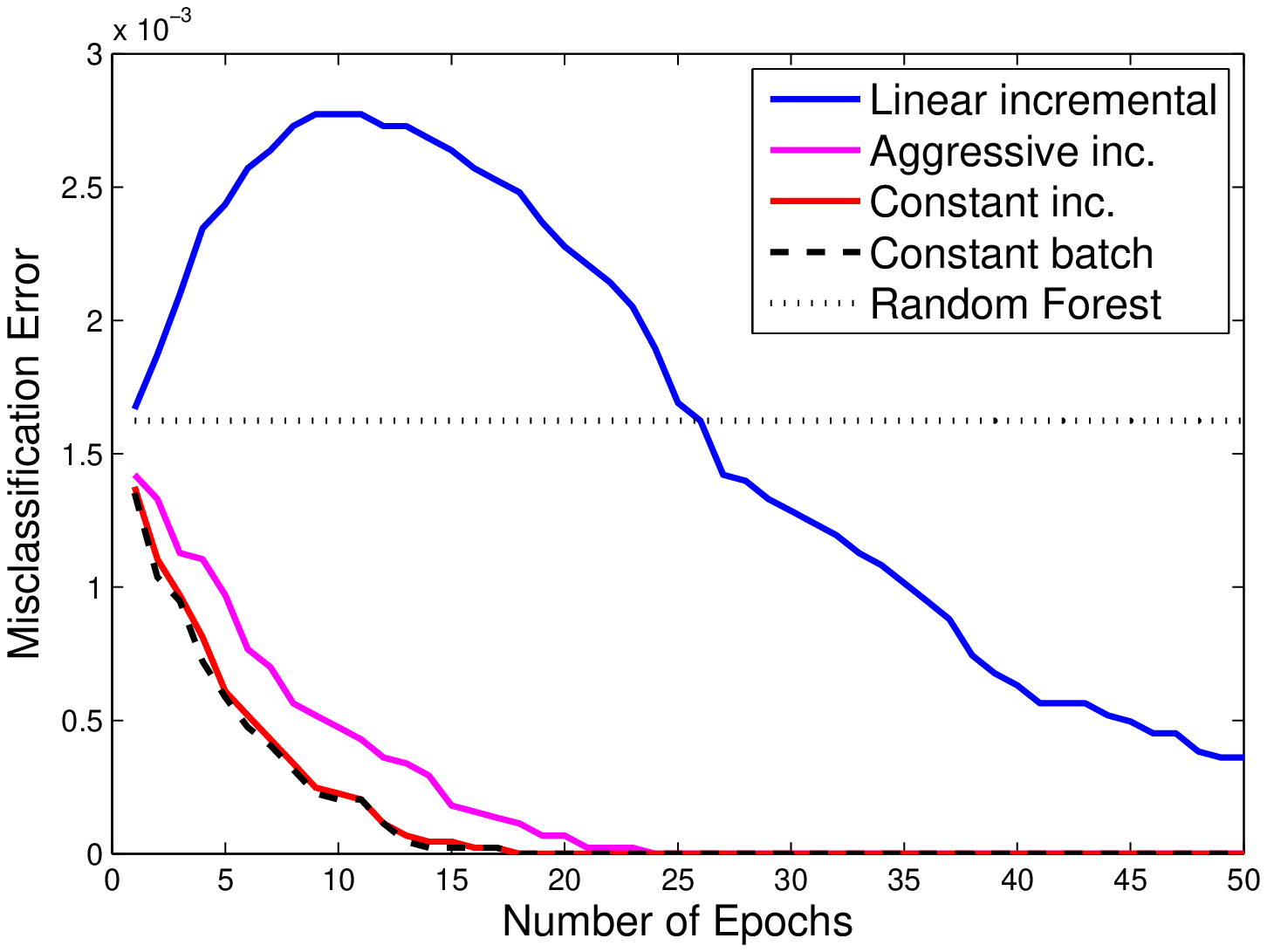}
\hspace{-1mm}\includegraphics[height=3.8cm]{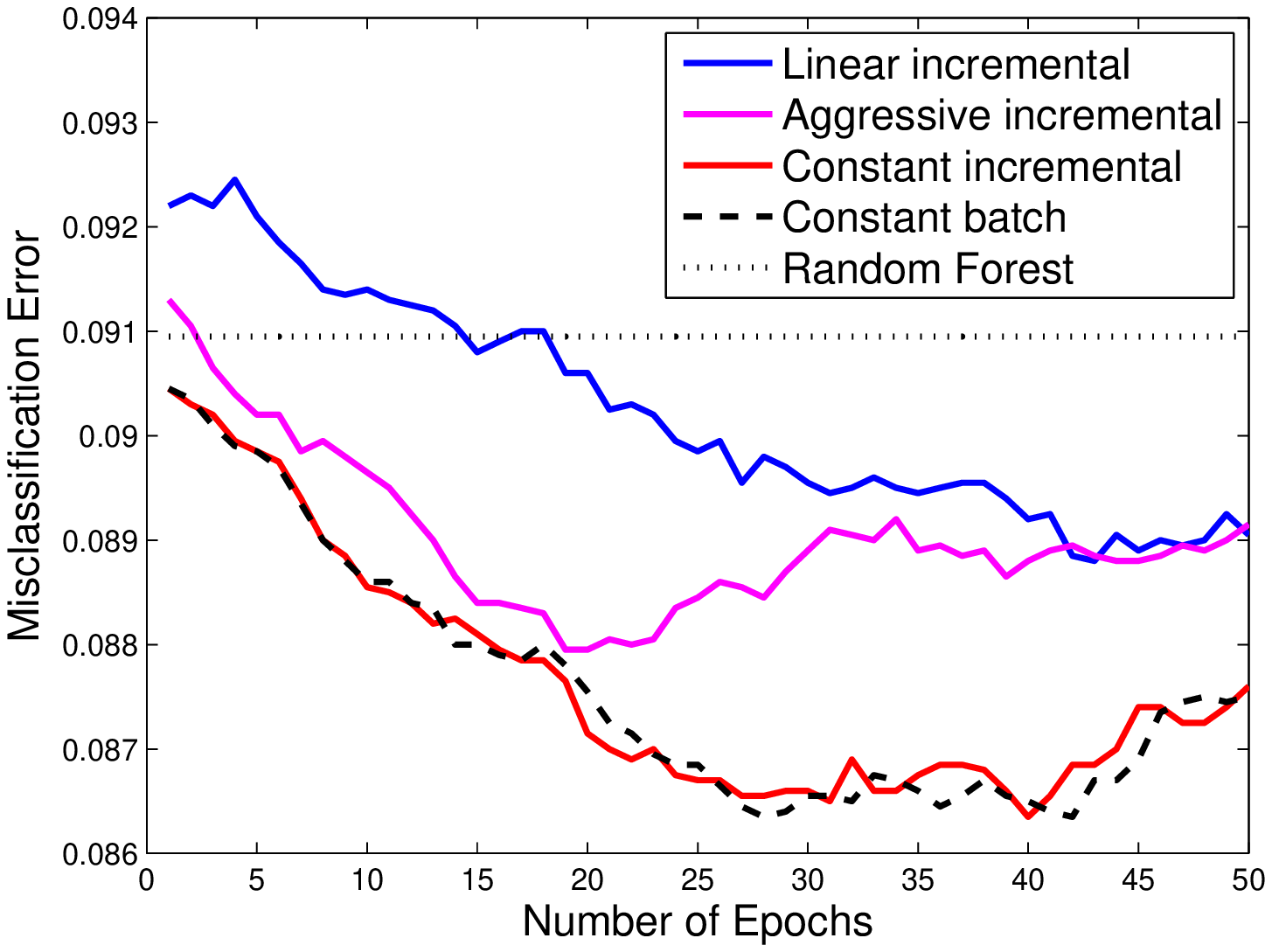}
\hspace{-1mm}\includegraphics[height=3.8cm]{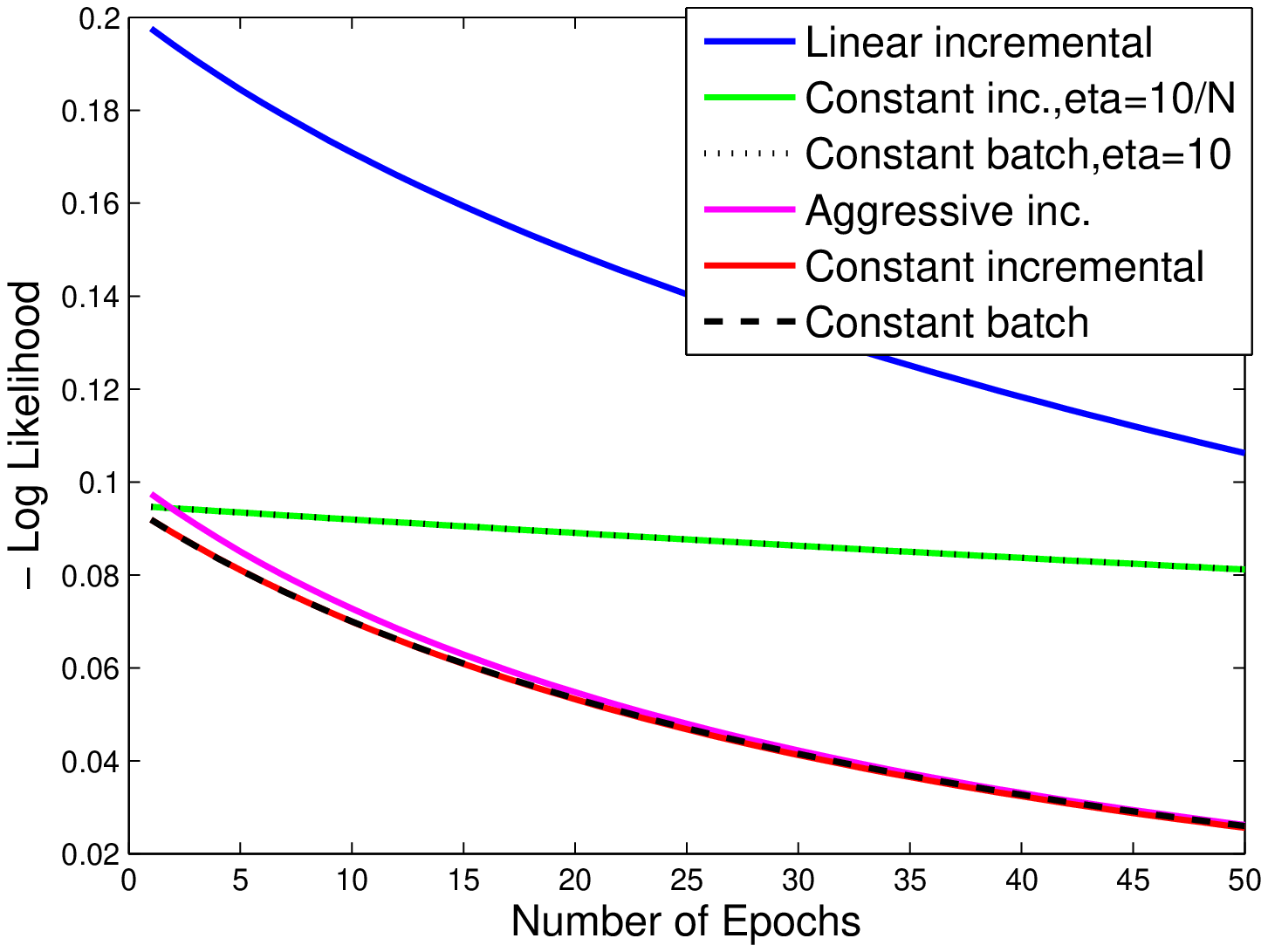}
\vskip -4mm
\caption{Experiments on the satimage  dataset for the incremental and batch market updates. Left: The training error vs. number of epochs.  Middle: The test error vs. number of epochs. Right: The negative log-likelihood function vs. number of training epochs. The learning rates are $\eta=100/N$ for the incremental update and $\eta=100$ for the batch update unless otherwise specified.}
\label{fig:SatLoss}
\end{figure}

In Figure \ref{fig:SatLoss} are plotted the misclassification errors on the training and test sets and the negative log-likelihood function vs. the number of training epochs, averaged over 10 runs. From Figure \ref{fig:SatLoss} one could see that the incremental and batch updates perform similarly in terms of the likelihood function, training and test errors. However, the incremental update is preferred since it is requires less memory and can handle an arbitrarily large amount of training data. The aggressive and constant markets achieve similar values of the negative log likelihood and similar training errors, but the aggressive market seems to overfit more since the test error is larger than the constant incremental ($p$-value$<0.05$). The linear market has worse values of the log-likelihood,  training and test errors ($p$-value$<0.05$).

\subsection{Evaluation of the Probability Estimation and Classification Accuracy on Synthetic Data}

We perform a series of experiments on synthetic datasets to evaluate the market's ability to predict class conditional probabilities $P(Y|{\bf x})$.
The experiments are performed on 5000 binary datasets with 50 levels of Bayes error
\vspace{-1mm}
\[
E=\int\hspace{-1mm} \min \{ p({\bf x},Y=0),p({\bf x},Y=1) \} d{\bf x},
\vspace{-1mm}
\]
ranging from 0.01 to 0.5 with equal increments. 
For each dataset, the two classes have equal frequency. Both $p({\bf x}|Y=k), k=0,1$ are normal distributions ${\cal N}(\mu_k,\sigma^2 I)$, with $\mu_0=0,\sigma^2=1$ and $\mu_1$ chosen in some random direction at such a distance to obtain the desired Bayes error.

For each of the 50 Bayes error levels, 100 datasets of size 200 were generated using the bisection method to find an appropriate $\mu_1$ in a random direction. Training of the participant budgets is done with $\eta=0.1$.

For each observation ${\bf x}$, the class conditional probability can be computed analytically using the Bayes rule
\vspace{-1mm}
\[
p^*(Y=1|{\bf x})=\frac{ p({\bf x}|Y=1) p(Y=1)}{p({\bf x},Y=0)+p({\bf x},Y=1)}
\vspace{-1mm}
\]
\begin{figure*}[htb]
\centering
\includegraphics[height=5.cm]{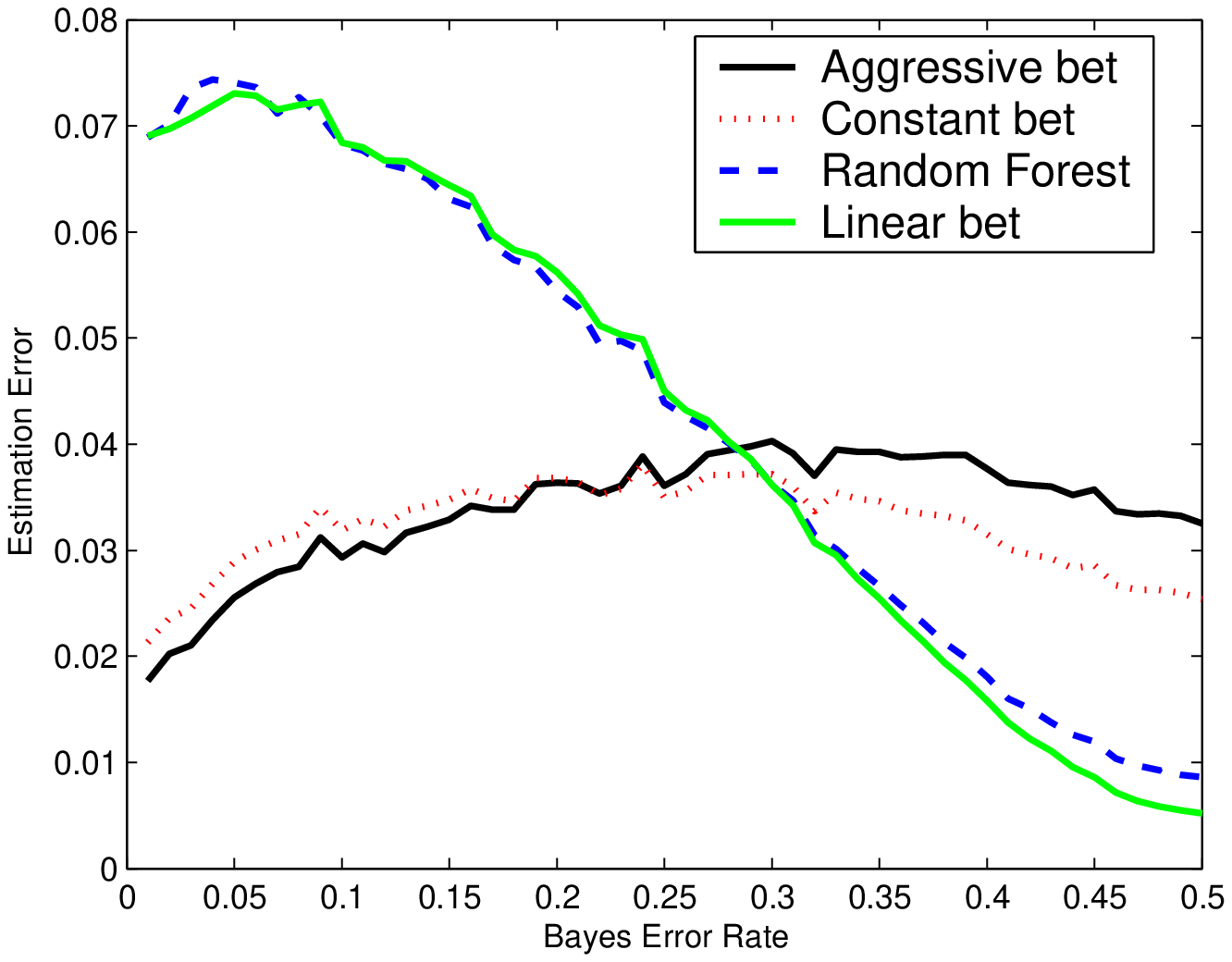}
\includegraphics[height=5.cm]{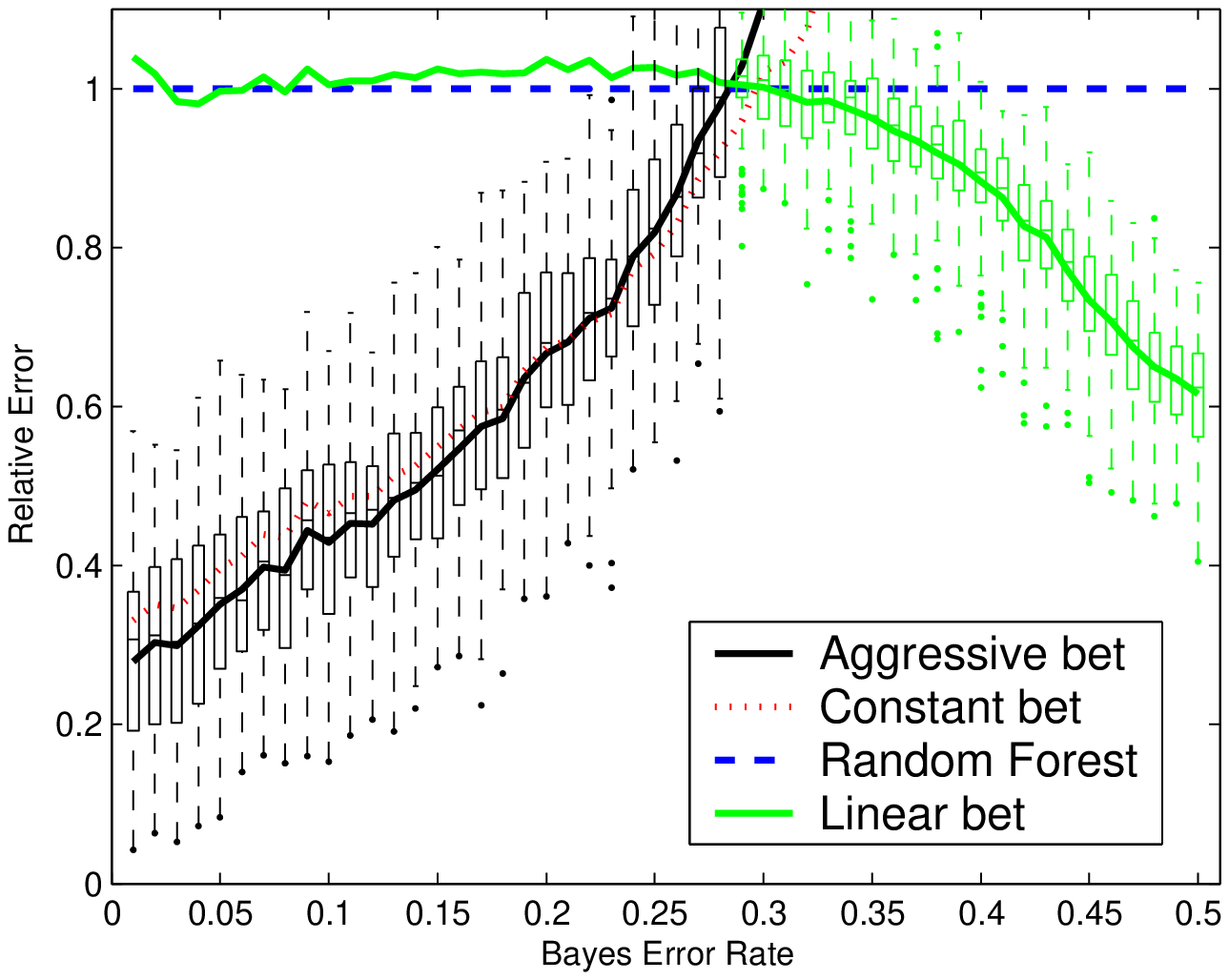}
\vskip -2mm
\caption{Left: Class probability estimation error vs problem difficulty for 5000 100D problems. Right: Probability estimation errors relative to random forest. The aggressive and linear betting are shown with box plots.}
\label{fig:dist100d}
\end{figure*}
\begin{figure*}[htb]
\centering
\includegraphics[height=5.cm]{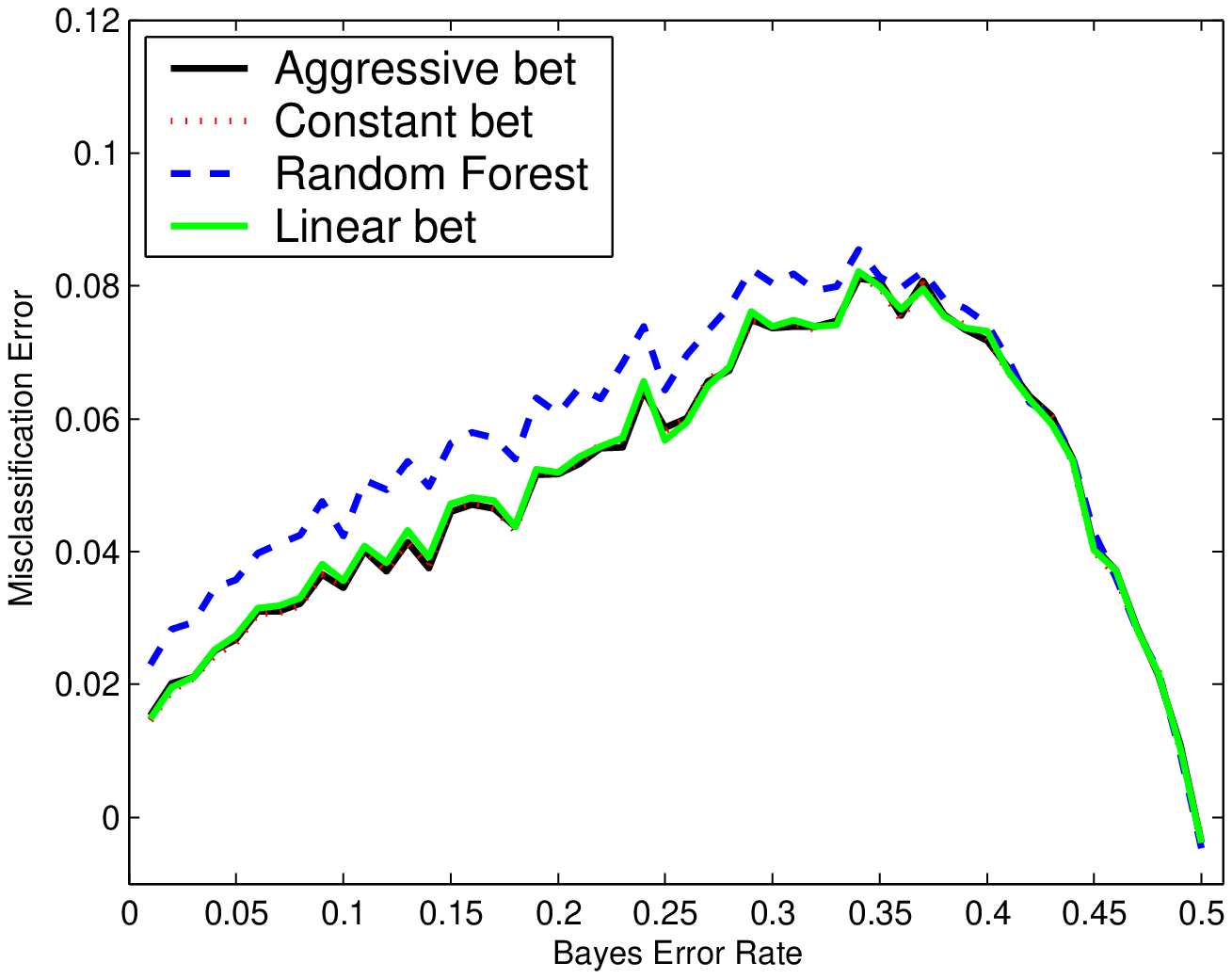}
\includegraphics[height=5.cm]{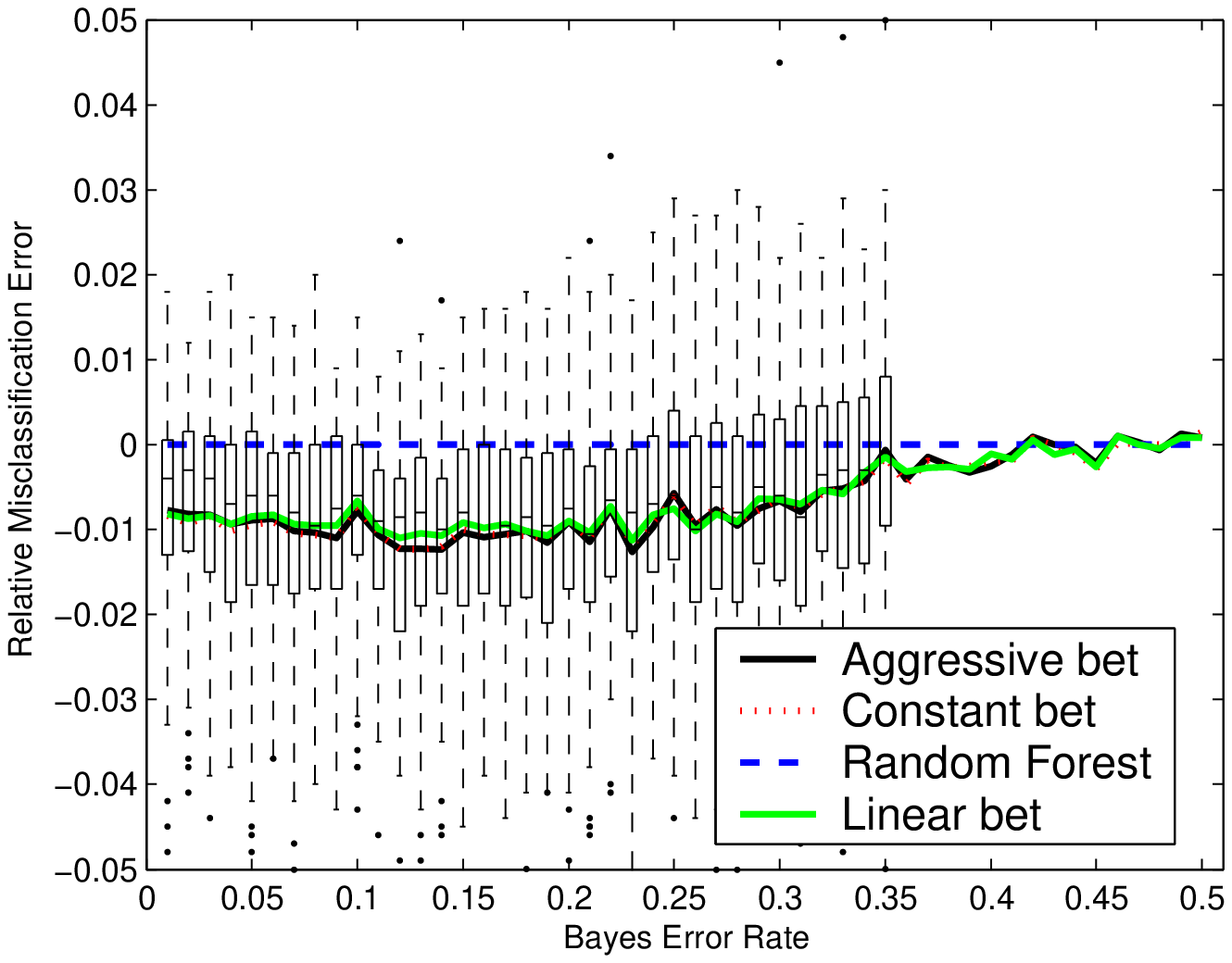}
\vskip -2mm
\caption{Left: Misclassification error minus Bayes error vs problem difficulty for 5000 100D problems. Right: Misclassification errors relative to random forest. The aggressive betting is shown with box plots.}
\label{fig:det100d}
\end{figure*}
An estimation $\hat p(y=1|{\bf x})$ obtained with one of the markets is compared to the true probability $p^*(Y=1|{\bf x})$ using the $L_2$ norm
\vspace{-1mm}
\[
E(\hat p, p^*)=\int (\hat{p}(y=1|{\bf x})-p^*(y=1|{\bf x}))^2 p({\bf x}) d{\bf x}
\vspace{-1mm}
\]
where $p({\bf x}) = p({\bf x},Y=0) + p({\bf x},Y=1)$.

In practice, this error is approximated using a sample of size 1000.  
The errors of the probability estimates obtained by the four markets are shown in Figure \ref{fig:dist100d} for a 100D problem setup. Also shown on the right are the errors relative to the random forest, obtained by dividing each error to the corresponding random forest error. As one could see, the aggressive and constant betting markets obtain significantly better ($p$-value $<0.01$) probability estimators than the random forest, for Bayes errors up to $0.28$. On the other hand, the linear betting market obtains probability estimators significantly better ($p$-value $<0.01$) than the random forest for Bayes error from $0.34$ to $0.5$. 

We also evaluated the misclassification errors of the four markets in predicting the correct class, for the same 5000 datasets. The difference between these misclassification errors and the Bayes error are shown in Figure \ref{fig:det100d}, left. The difference between these misclassification errors and the random forest error are shown in Figure \ref{fig:det100d}, right. We see that all markets with trained participants predict significantly better ($p$-value $<0.01$) than random forest for Bayes errors up to $0.3$, and behave similar to random forest for the remaining datasets.

\subsection{Comparison with Random Forest on UCI Datasets} \label{sec:expRF}

In this section we conduct an evaluation on 31 datasets from the UCI machine learning repository \citep{blake1998uci}. 
The optimal number of training epochs and $\eta$ are meta-parameters that need to be chosen appropriately for each dataset. 
We observed experimentally that $\eta$ can take any value up to a maximum that depends on the dataset. In these experiments we took $\eta=10/N_{train}$. The best number of epochs was chosen by ten fold cross-validation.

\begin{table*} 
\small
\begin{center}
\caption{The misclassification errors for 31 datasets from the UC Irvine Repository are shown in percent ($\%$).\label{tab:uci}. The markets evaluated are our implementation of random forest (RF), and markets with Constant (CB), Linear (LB) and respectively Aggressive (AB) Betting.  RFB contains the random forest results from \citep{breiman_random_2001}.}
\input{tableSplits}
\normalsize
\end{center}
\end{table*}

In order to compare with the results in \citep{breiman_random_2001}, the training and test sets were randomly subsampled from the available data, with $90\%$ for training and $10\%$ for testing. The exceptions are the \verb satimage , \verb zipcode , \verb hill-valley  and \verb poker datasets with test sets of size $2000, 2007, 606, 10^6$ respectively.  All results were averaged over 100 runs.

We present two random forest results. 
In the column named RFB are presented the random forest results  from \citep{breiman_random_2001}where each tree node is split based on a random feature. In the column named RF we present the results of our own RF implementation with splits based on random features. The leaf nodes of the random trees from our RF implementation are used as specialized participants for all the markets evaluated. 

The CB, LB and AB columns are the performances of the constant, linear and respectively aggressive markets on these datasets.

Significant mean differences ($\alpha<0.01$) from RFB are shown with $+,-$ for when RFB is worse respectively better. 
Significant paired $t$-tests \citep{demsar2006statistical} ($\alpha<0.01$) that compare the markets with our RF implementation are shown with $\bullet,\dag$ for when RF is worse respectively better. 

The constant, linear and aggressive markets significantly outperformed our RF implementation on 22, 19 respectively 22 datasets out of the 31 evaluated. They were not significantly outperformed by our RF implementation on any of the 31 datasets.

Compared to the RF results from \cite{breiman_random_2001} (RFB), CB, LB and AB significantly outperformed RFB on 6,5,6 datasets respectively, and were not significantly outperformed on any dataset.


\subsection{Comparison with Implicit Online Learning on UCI Datasets}

We implemented the implicit online learning \citep{kulis2010implicit} algorithm for classification with linear aggregation. The objective of implicit online learning is to minimize the loss $\ell(\beta)$ in a \textit{conservative} way.  The \textit{conservativeness} of the update is determined by a Bregman divergence
\[
	D(\beta,\beta^t) = \phi(\beta) - \phi(\beta^t) - \langle \nabla \phi(\beta^t), \beta-\beta^t \rangle
\]
where $\phi(\beta)$ are real-valued strictly convex functions.  Rather than minimize the loss function itself, the function
\[
	f_t(\beta) = D(\beta,\beta^t) + \eta_t \ell(\beta)
\]
is minimized instead. Here $\eta_t$ is the learning rate. The Bregman divergence ensures that the optimal $\beta$ is not \textit{too far} from $\beta^t$.  The algorithm for implicit online learning is as follows
\[
\begin{split}
	\tilde{\beta}^{t+1} &= \argmin_{\beta \in \mathbb{R}^M} f_t(\beta) \\
	\beta^{t+1} &= \argmin_{\beta \in S} D(\beta,\tilde{\beta}^{t+1})
\end{split}
\]
The first step solves the unconstrained version of the problem while the second step finds the \textit{nearest} feasible solution to the unconstrained minimizer subject to the Bregman divergence. 

For our problem we use
\[
	\ell(\beta) = -\log(c_y(\beta))
\]
where $c_y(\beta)$ is the constant market equilibrium price for ground truth label $y$. We chose the squared Euclidean distance $D(\beta,\beta^t) = \| \beta - \beta^t \|_2^2$ as our Bregman divergence and learning rate $\eta_t = 1/\sqrt{t}$. To ensure that $c = \sum_{m=1}^M h_m \beta_m = H \beta$ is a valid probability vector, the feasible solution set is therefore $S = \{ \beta \in [0,1]^M\ :\ \sum_{m=1}^M \beta_m = 1 \}$.  This gives the following update scheme
\[
\begin{split}
	\tilde{\beta}^{t+1} &= \beta^t + \eta_t \frac{1}{p} (H^y)^T \\
	\beta^{t+1} &= \argmin_{\beta \in S} \left \{ \|\beta - \tilde{\beta}^{t+1} \|_2^2 \right \}
\end{split}
\]
where $	H^y = \left ( h_1^y,\ h_2^y,\ \hdots,\ h_M^y \right )$ is the vector of classifier outputs for the true label $y$, $q = H^y \beta^t,\ r = H^y (H^y)^T$ and $p = \frac{1}{2} \left ( q + \sqrt{q^2 + 4 \eta_t r} \right )$.

\begin{table*} 
\small
\begin{center}
\caption{Comparison with Implicit Online Learning and random forest using 10-fold cross-validation.\label{tab:kulis}}
\input{tableKulis}
\normalsize
\end{center}
\end{table*}

The results presented in Table \ref{tab:kulis} are obtained by 10 fold cross-validation. The cross-validation errors were averaged over 10 different permutations of the data in the cross-validation folds. 

The results from CB online and implicit online are obtained in one epoch. The results from the CB offline and implicit offline columns are obtained in an off-line fashion using an appropriate number of epochs (up to 10) to obtain the smallest cross-validated error on a random permutation of the data that is different from the 10 permutations used to obtain the results.

The comparisons are done with paired $t$-tests and shown with \textasteriskcentered\; and \ddag \;when the
constant betting market is significantly ($\alpha<0.01$) better or worse than the corresponding implicit online learning. 
We also performed a comparison with our RF implementation, and significant differences are shown with {\textbullet}\; and \dag. 

Compared to RF, implicit online learning won 5-0, CB online won in 9-1 and  CB offline won 12-0.

Compared to implicit online, which performed identical with implicit offline, both CB online and CB offline won 9-0.

\subsection{Comparison with Adaboost for Lymph Node Detection}

Finally, we compared the linear aggregation capability of the artificial prediction market with adaboost for a lymph node detection problem. The system is setup as described in \cite{barbu2012automatic}, namely a set of lymph node candidate positions $(x,y,z)$ are obtained using a trained detector. Each candidate is segmented using gradient descent optimization and about 17000 features are extracted from the segmentation result. Using these features, adaboost constructed 32 weak classifiers. Each weak classifier is associated with one feature, splits the feature range into 64 bins and returns a predefined value ($1$ or $-1$), for each bin. 

Thus, one can consider there are $M=32\times 64=2048$ specialized participants, each betting for one class ($1$ or $-1$) for any observation that falls in its domain. The participants are given budgets $\beta_{ij},i=1,..,32,j=1,..,64$ where $i$ is the feature index and $j$ is the bin index.
The participant budgets $\beta_{ij},j=1,...,64$ corresponding to the same feature $i$ are initialized the same value $\beta_i$, namely the adaboost coefficient. For each bin, the return class $1$ or $-1$ is the outcome for which the participant will bet its budget.

The constant betting market of the 2048 participants is initialized with these budgets and trained with the same training examples that were used to train the adaboost classifier. 

The obtained constant market probability for an observation ${\bf x}=(x_1,...,x_{32})$ is based on the bin indexes ${\bf b}=(b_1(x_1),...,b_{32}(x_{32})$:
\begin{equation}
p(y=1|{\bf b})=\frac{\sum_{i=1}^{32} \beta_{i,b_i} h_i(b_i)}{\sum_{i=1}^{32} \beta_{i,b_i}}
\end{equation}

An important issue is that the number $N_{pos}$ of positive examples is much smaller than the number $N_{neg}$ of negatives. Similar to adaboost, the sum of the weights of the positive examples should be the same as the sum of weights of the negatives. To accomplish this in the market, we use the weighted update rule Eq. \eqref{eq:wtupdate}, with $w_{pos}=\frac{1}{N_{pos}}$ for each positive example and $w_{neg}=\frac{1}{N_{neg}}$ for each negative.

\begin{figure}[htb]
\centering
\includegraphics[height=5.5cm]{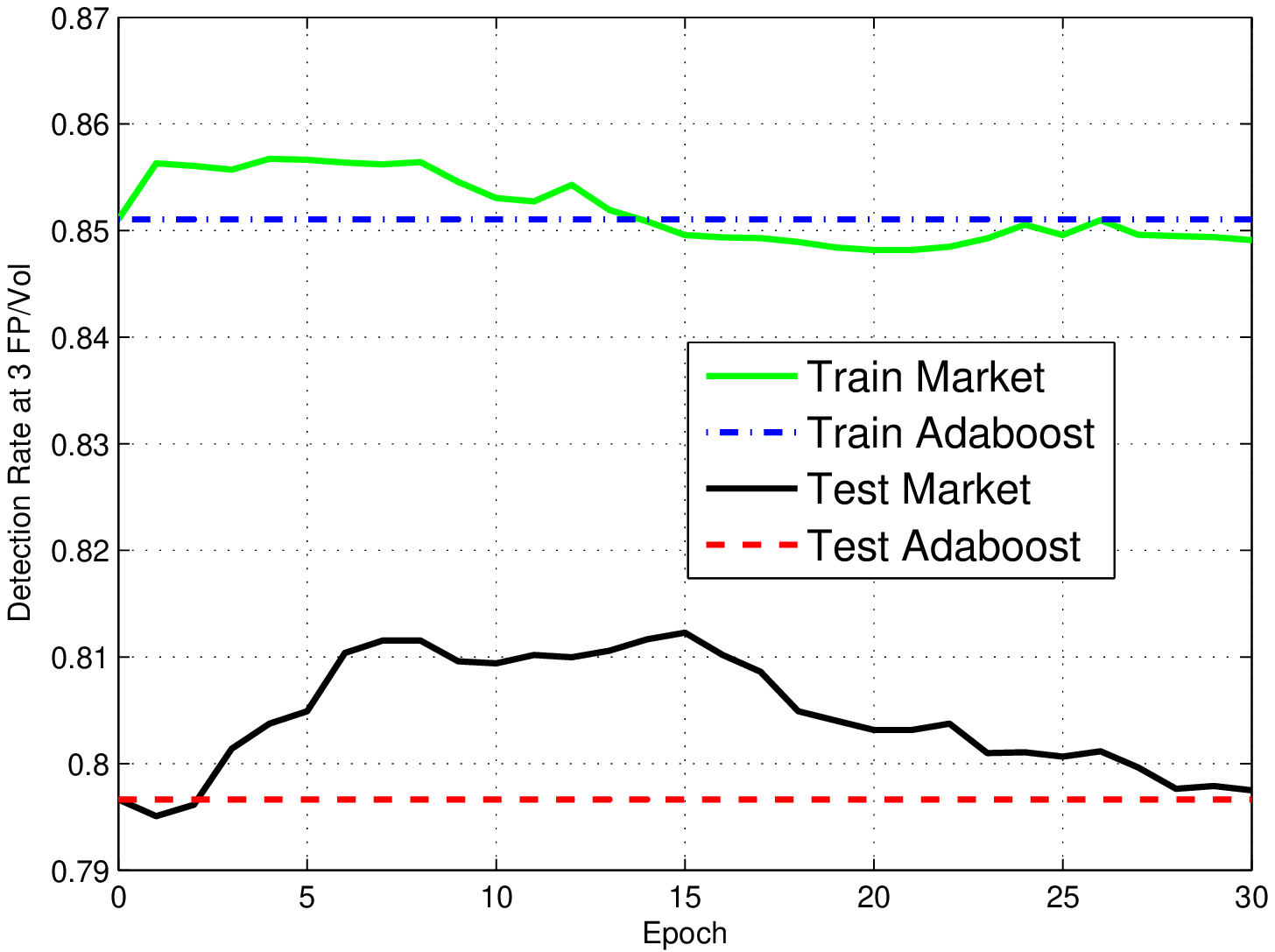}
\includegraphics[height=5.5cm]{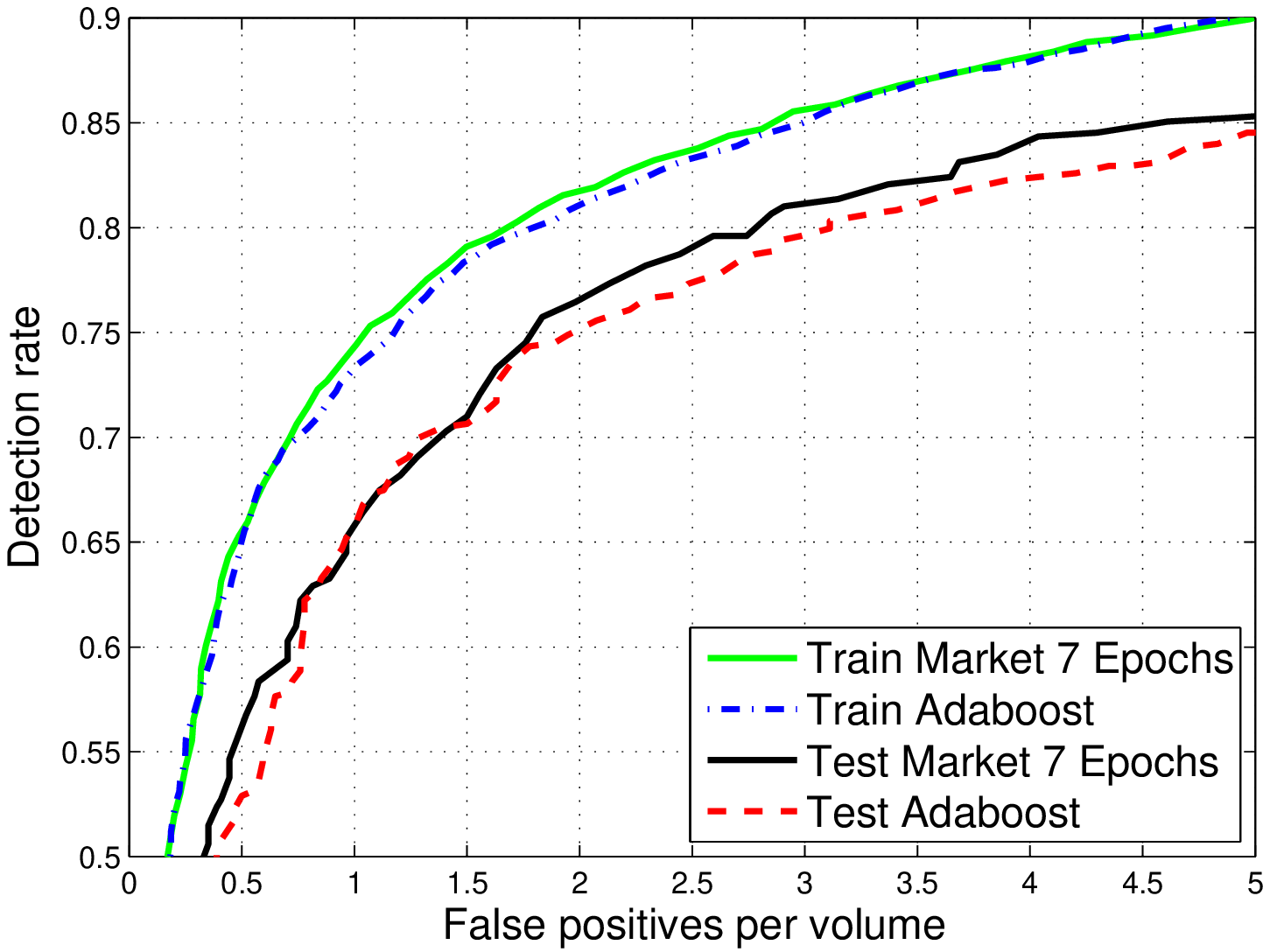}
\vskip -2mm
\caption{Left: Detection rate at 3 FP/vol vs. number of training epochs for a lymph node detection problem. Right: ROC curves for adaboost and the constant betting market with participants as the 2048 adaboost weak classifier bins. The results are obtained with six-fold cross-validation.}
\label{fig:lnroc}
\end{figure}

The adaboost classifier and the constant market were evaluated for a lymph node detection application on a dataset containing 54 CT scans of the pelvic and abdominal region, with a total of 569 lymph nodes, with six-fold cross-validation. The evaluation criterion is the same for all methods, as specified in \cite{barbu2012automatic}. A lymph node detection is considered correct if its center is inside a manual solid lymph node segmentation and is incorrect if it not inside any lymph node segmentation (solid or non-solid).

In Figure \ref{fig:lnroc}, left, is shown the training and testing detection rate at 3 false positives per volume (a clinically acceptable false positive rate) vs the number of training epochs. We see the detection rate increases to about $81\%$ for epochs 6 to 16 epochs and then gradually decreases. In Figure \ref{fig:lnroc}, right, are shown the training and test ROC curves of adaboost and the constant market trained with 7 epochs. In this case the detection rate at 3 false positives per volume improved from $79.6\%$ for adaboost to $81.2\%$ for the constant market. The $p$-value for this difference was 0.0276 based on paired $t$-test. 

\section{Conclusion and Future Work}

This paper presents a theory for artificial prediction markets for the purpose of supervised learning of class conditional probability estimators. The artificial prediction market is a novel online learning algorithm that can be easily implemented for two class and multi class applications. Linear aggregation, logistic regression as well as certain kernel methods can be viewed as particular instances of the artificial prediction markets. Inspired from real life, specialized classifiers that only bet on subsets of the instance space $\Omega$ were introduced. Experimental comparisons on real and synthetic data show that the prediction market usually outperforms random forest, adaboost and implicit online learning in prediction accuracy. 

The artificial prediction market shows the following promising features:
\begin{enumerate}
	\item It can be {\em updated online} with minimal computational cost when a new observation $({\bf x},y)$ is presented.
	\item It has a simple form of the update iteration that can be easily implemented.
	\item For multi-class classification it can {\em fuse information} from all types of binary or multi-class classifiers: e.g. trained {\em one-vs-all, many-vs-many}, multi-class decision tree, etc.
	\item It can obtain meaningful probability estimates when only a subset of the market participants are involved for a particular instance ${\bf x}\in X$. This feature is useful for learning on manifolds \citep{belkin2004ssl,elgammal2004ibp,saul2003tgf}, where the location on the manifold decides which market participants should be involved. For example, in face detection, different face part classifiers (eyes, mouth, ears, nose, hair, etc) can be involved in the market, depending on the orientation of the head hypothesis being evaluated. 
	\item Because of their betting functions, the {\em specialized} market participants can decide for which instances they bet and how much. This is another way to combine classifiers, different from the boosting approach where all classifiers participate in estimating the class probability for each observation.
\end{enumerate}
We are currently extending the artificial prediction market framework to regression and density estimation. These extensions involve contracts for uncountably many outcomes but the update and the market price equations extend naturally. 

Future work includes finding explicit bounds for the generalization error based on the number of training examples. Another item of future work is finding other generic types specialized participants that are not leaves of random or adaboost trees. For example, by clustering the instances ${\bf x}\in \Omega$, one could find regions of the instance space $\Omega$ where simple classifiers (e.g. logistic regression, or betting for a single class) can be used as specialized market participants for that region. 


 \section*{Acknowledgments}

The authors wish to thank Jan Hendrik Schmidt from Innovation Park Gmbh. for stirring in us the excitement for the prediction markets. The authors acknowledge partial support from FSU startup grant and ONR N00014-09-1-0664.
\bibliographystyle{natbib}
{\small 
\bibliography{predmarket}
}

\section*{Appendix: Proofs} 

\begin{proof}[of Theorem \ref{thm:budget}]
From eq. \eqref{eq:budgetupdate}, the total budget $\sum_{m=1}^M \beta_m$ is conserved if and only if
\begin{equation}
	\sum_{m=1}^M\sum_{k=1}^K \beta_m\phi_m^k({\bf x},{\bf c})=\sum_{m=1}^M \beta_m\phi_m^y({\bf x},{\bf c})/c_y \label{eq:bugcons}
\end{equation}
Denoting $n=\sum_{m=1}^M\sum_{k=1}^K \beta_m\phi_m^k({\bf x},{\bf c})$, and since the above equation must hold for all $y$, we obtain that 
eq. \eqref{eq:budgetcons} is a necessary condition and also $c_k\not =0, k=1,...,K$, which means $c_k>0, k=1,...,K$.
Reciprocally, if $c_k>0$ and eq. \eqref{eq:budgetcons} hold for all $k$, dividing by $c_k$ we obtain eq. \eqref{eq:bugcons}.

\end{proof}

\begin{proof}[of Remark \ref{rem:unique}] Since the total budget is conserved and is positive, there exists a $\beta_m>0$, therefore $\sum_{m=1}^M\beta_m\phi_m^k({\bf x},0)>0$, which implies $\lim_{c_k\to 0} f_k(c_k)=\infty$. From the fact that $f_k(c_k)$ is continuous and strictly decreasing, with $\lim_{c_k\to 0} f_k(c_k)=\infty$ and  $\lim_{c_k\to 1} f_k(c_k)=0$, it implies that for every $n> 0$ there exists a unique $c_k$ that satisfies $f_k(c_k)=n$. 
\end{proof}

\begin{proof}[of Theorem \ref{thm:monbet}] From Remark \ref{rem:unique} we get that for every $n\geq n_k, n>0$ there is a unique $c_k(n)$ such that $f_k(c_k(n))=n$. Moreover, following the proof of Remark \ref{rem:unique} we see that $c_k(n)$ is continuous and strictly decreasing on $(n_k,\infty)$, with  $\lim_{n\to \infty}c_k(n)=0$. 

If  $\max_k n_k>0$, take $n^*=\max_k n_k$. There exists $k\in \{1, ..., K\} $ such that $n_k=n^*$, so $c_k(n^*)=1$, therefore $\sum_{j=1}^K c_j(n^*)\geq 1$.

If  $\max_k n_k=0$ then $n_k=0, k=1,...,K$ which means $\phi_m^k({\bf x},1)=0,k=1,...,K$ for all $m$ with $\beta_m>0$. Let  $a^k_m=\min \{c| \phi_m^k({\bf x},c)=0\}$. We have  $a_m^k>0$ for all $k$ since $\phi_m^k({\bf x},0)>0$.
Thus $\lim _{n\to 0_+}c_k(n)=\max_m a_m^k\geq a_1^k$, where we assumed that $\phi_1({\bf x},{\bf c})$ satisfies Assumption \ref{asm:total}. But  from Assumption  \ref{asm:total} there exists $k$ such that $a_1^k=1$.  Thus  $\lim _{n\to 0_+}\sum_{k=1}^K c_k(n)\geq \sum_{k=1}^K a_1^k>1$ so there exists $n^*$ such that $\sum_{k=1}^K c_k(n^*)\geq 1$.

Either way, since $\sum_{k=1}^K c_k(n)$ is continuous, strictly decreasing, and since  $\sum_{k=1}^K c_k(n^*)\geq 1$ and  $\lim _{n\to \infty}\sum_{k=1}^K c_k(n)=0$, there exists a unique $n>0$ such that $\sum_{k=1}^K c_k(n)=1$. For this $n$, from Theorem \ref{thm:budget} follows that the total budget is conserved for the price ${\bf c}=(c_1(n),...,c_K(n))$. Uniqueness follows from the uniqueness of $c_k(n)$ and the uniqueness of $n$. 
\end{proof}

\begin{proof}[of Theorem \ref{thm:ctbet}] The price equations \eqref{eq:budgetcons} become:
\[
\sum_{m=1}^M \beta_m\phi_m^k({\bf x})=c_k\sum_{k=1}^K \sum_{m=1}^M\beta_m\phi_m^k({\bf x}), \quad \forall k=1,...,K. 
\]
which give the result from eq. \eqref{eq:ctbet}.

If  $\phi_m^k({\bf x})=\eta h_m^k({\bf x})$, using  $\sum_{k=1}^K h_m^k({\bf x})=1$, the denominator of eq. \eqref{eq:ctbet} becomes  
\[
 \sum_{k=1}^K\sum_{m=1}^M \beta_m \phi_m^k({\bf x})=\eta  \sum_{m=1}^M \beta_m\sum_{k=1}^K h_m^k({\bf x})=\eta \sum_{m=1}^M \beta_m
\]
so
\[
c_k=\frac{\eta\sum_{m=1}^M \beta_mh_m^k({\bf x})}{\eta\sum_{m=1}^M \beta_m}=\sum_m \alpha_m h_m^k({\bf x}), \quad \forall k=1,...,K
\]
\end{proof}

\begin{proof}[of Theorem \ref{thm:ctloss}]
For the current parameters $\gamma=(\gamma_1,...,\gamma_M)=(\sqrt{\beta_1},...,\sqrt{\beta_m})$ and an observation $({\bf x}_i,y_i)$, we have the market price for label $y_i$:
\vspace{-1mm}
\begin{equation}
c_{y_i}({\bf x}_i)=	\sum_{m=1}^M \gamma_m^2 \phi^{y_i}_m({\bf x}_i)/(\sum_{m=1}^M \sum_{k=1}^K\gamma_m^2 \phi^{k}_m({\bf x}_i))\label{eq:ctprice}
\vspace{-3mm}
\end{equation}
So the log-likelihood is
\begin{equation}
L(\gamma)=\frac{1}{N}\sum_{i=1}^N\log c_{y_i}({\bf x}_i)=\frac{1}{N}\sum_{i=1}^N\log \sum_{m=1}^M \gamma_m^2 \phi^{y_i}_m({\bf x}_i)-\frac{1}{N}\sum_{i=1}^N\log \sum_{m=1}^M \sum_{k=1}^K \gamma_m^2 \phi^{k}_m({\bf x}_i)\label{eq:ctlogc}
\end{equation}
We obtain the gradient components:
\vspace{-2mm}
\begin{equation}
	\frac{\partial L(\gamma)}{\partial \gamma_j}= \frac{1}{N}\sum_{i=1}^N\left(\frac{\gamma_j\phi^{y_i}_j({\bf x}_i)}{\sum_{m=1}^M \gamma_m^2 \phi^{y_i}_m({\bf x}_i)}-\frac{\gamma_j\sum_{k=1}^K \phi^{k}_j({\bf x}_i)}{\sum_{m=1}^M  \sum_{k=1}^K\gamma_m^2 \phi^{k}_m({\bf x}_i)}\right)\label{eq:ctgradupd}
\vspace{-1mm}
\end{equation}
Then from \eqref{eq:ctprice} we have $\sum_{m=1}^M \gamma_m^2 \phi^{y_i}_m({\bf x}_i)=B({\bf x}_i)c_{y_i}({\bf x}_i)$. Hence \eqref{eq:ctgradupd} becomes
\vspace{-2mm}
\[
	\frac{\partial L(\gamma)}{\partial \gamma_j}=\frac{\gamma_j}{N}\sum_{i=1}^N\frac{1}{B({\bf x}_i)}\left(\frac{\phi^{y_i}_j({\bf x}_i)}{c_{y_i}({\bf x}_i)}-\sum_{k=1}^K \phi_j^k({\bf x}_i)\right).
\vspace{-1mm}
\]
Write $u_j=\frac{1}{N}\sum_{i=1}^N\frac{1}{B({\bf x}_i)}\left(\frac{\phi^{y_i}_j({\bf x}_i)}{c_{y_i}({\bf x}_i)}-\sum_{k=1}^K \phi_j^k({\bf x}_i)\right)$, then $\frac{\partial L(\gamma)}{\partial \gamma_j}=\gamma_ju_j$.
The batch update \eqref{eq:ctbatchupd} is $\beta_j\leftarrow \beta_j +\eta\beta_ju_j$. By taking the square root we get the update in $\gamma$
\[
\gamma_j\leftarrow \gamma_j\sqrt{1+\eta u_j}=\gamma_j+\gamma_j(\sqrt{1+\eta u_j}-1)=\gamma_j+\gamma_j\frac{\eta u_j}{\sqrt{1+\eta u_j}+1}=\gamma_j'.
\]
We can write the Taylor expansion:
\[
L(\gamma')=L(\gamma)+(\gamma'-\gamma)^T\nabla L(\gamma)+\frac{1}{2}(\gamma'-\gamma)^T H(L)(\zeta)(\gamma'-\gamma)
\]
so
\[
\begin{split}
L(\gamma')&=L(\gamma)+\sum_{j=1}^M \gamma_j u_j\frac{\eta\gamma_ju_j}{\sqrt{1+\eta u_j}+1}+\eta^2A(\eta)
=L(\gamma)+\eta\sum_{j=1}^M \frac{\gamma_j^2u_j^2}{\sqrt{1+\eta u_j}+1}+\eta^2A(\eta)
\end{split}
\]
where $|A(\eta)|$ is bounded in a neighborhood of $0$.

Now assume that $\nabla L(\gamma)\not =0$, thus $\gamma_ju_j\not =0$ for some $j$. Then $\sum_{j=1}^M \frac{\gamma_j^2u_j^2}{\sqrt{1+\eta u_j}+1}>0$ hence $L(\gamma')>L(\gamma)$ for any $\eta$ small enough.

Thus as long as $\nabla L(\gamma)\not =0$ the batch update \eqref{eq:ctbatchupd} with any $\eta$ sufficiently small will increase the likelihood function.

The batch update \eqref{eq:ctbatchupd} can be split into $N$ per-observation updates of the form \eqref{eq:ctincupd}.
\end{proof}
\end{document}

%% file: notation.tex
\newcommand{\argmin}{\operatornamewithlimits{argmin}}

%% file: tableSplits.tex
\begin{tabular} {|l|c|c|c|c|c|c|c|c|c|} 
\hline
Data &$N_{\text{train} } $ &$N_{\text{test} } $ &$F$ &$K$ &RFB &RF &CB &LB &AB \\
\hline 
breast-cancer &683 &-- &9 &2 &2.7 &{ 2.5}  &{ 2.4}  &{ 2.4}  &{ 2.4}  \\
sonar &208 &-- &60 &2  &18.0 &{16.6}  &{14.1} \textbullet+ &{14.2} \textbullet+&{14.1} \textbullet+\\
vowel &990 &-- &10 &11  &3.3 &{ 2.9}  &{ 2.6} \textbullet+ &{ 2.7} + &{ 2.6} \textbullet+ \\
ecoli &336 &-- &7 &8  &13.0 &{ 12.9}  &{ 12.9}  &{ 12.8}  &{ 12.9}  \\
german &1000 &-- &24 &2 &26.2 &{ 25.5}  &{ 24.9} \textbullet+ &{ 25.1} &{ 24.9} \textbullet+ \\
glass &214 &-- &9 &6 &21.2 &{23.5}  &{22.2} \textbullet &{22.4}  &{22.2} \textbullet \\
image &2310 &-- &19 &7  &2.7 &{ 2.7}  &{ 2.5} \textbullet &{ 2.5} \textbullet &{ 2.5} \textbullet \\
ionosphere &351 &-- &34 &2  &7.5 &{7.4}  &{6.7} \textbullet &{6.9} \textbullet &{6.7} \textbullet \\
letter-recognition &20000 &-- &16 &26  &4.7 &{ 4.2} + &{ 4.2} \textbullet+ &{ 4.2} \textbullet+ &{ 4.2} \textbullet+ \\
liver-disorders &345 &-- &6 &2  &24.7 &{ 26.5}  &{ 26.3}  &{ 26.2}  &{ 26.2}  \\
pima-diabetes &768 &-- &8 &2  &24.3 &{ 24.1}  &{ 23.8}  &{ 23.7}  &{ 23.8}  \\
satimage &4435 &2000 &36 &6  &10.5 &{ 10.1} + &{ 10.0} \textbullet+ &{ 10.1} \textbullet+ &{ 10.0} \textbullet+ \\
vehicle &846 &-- &18 &4  &26.4 &{ 26.3}  &{ 26.1}  &{ 26.2}  &{ 26.1}  \\
voting-records &232 &-- &16 &2 &4.6 &{5.3}  &{4.2} \textbullet &{4.2} \textbullet &{4.2} \textbullet \\
zipcode &7291 &2007 &256 &10  &7.8 &{ 7.7}  &{ 7.6} \textbullet+ &{ 7.7} \textbullet+ &{ 7.6} \textbullet+ \\
abalone &4177 &-- &8 &3 &--  &{45.5}  &{45.4}  &{45.4}  &{45.4}  \\
balance-scale &625 &-- &4 &3  &-- &{15.4}  &{15.4}  &{15.4}  &{15.4}  \\
car &1728 &-- &6 &4 &--  &{2.8}  &{2.0} \textbullet &{2.2} \textbullet &{2.0} \textbullet \\
connect-4 &67557 &-- &42 &3  &-- &{19.6}  &{19.3} \textbullet &{19.4} \textbullet &{19.5} \textbullet \\
cylinder-bands &277 &-- &33 &2  &-- &{22.7}  &{20.9} \textbullet &{21.1} \textbullet &{20.9} \textbullet \\
hill-valley &606 &606 &100 &2 &-- &{46.9}  &{45.8} \textbullet &{46.3} \textbullet &{45.8} \textbullet \\
isolet &1559 &-- &617 &26 &--  &{17.0}  &{15.7} \textbullet &{15.8} \textbullet &{15.7} \textbullet \\
king-rook-vs-king &28056 &-- &6 &18  &-- &{15.6}  &{15.4} \textbullet &{15.4} \textbullet &{15.4} \textbullet \\
king-rk-vs-k-pawn &3196 &-- &36 &2  &-- &{2.0}  &{1.5} \textbullet &{1.6} \textbullet &{1.5} \textbullet \\
madelon &2000 &-- &500 &2 &--  &{46.1}  &{45.2} \textbullet &{45.3} \textbullet &{45.2} \textbullet \\
magic &19020 &-- &10 &2 &--  &{12.0}  &{11.9} \textbullet &{11.9} \textbullet &{11.9} \textbullet \\
musk &6598 &-- &166 &2 &--  &{3.7}  &{3.5} \textbullet &{3.6} \textbullet &{3.5} \textbullet \\
poker &25010 &$10^6$ &10 &10  &-- &{43.2}  &{43.1} \textbullet &{43.1} \textbullet &{43.1} \textbullet \\
SAheart &462 &-- &9 &2 &--  &{30.8}  &{30.8}  &{30.7}  &{30.8}  \\
splice-junction &3190 &-- &59 &3  &-- &{18.9}  &{17.7} \textbullet &{18.2} \textbullet &{17.7} \textbullet \\
yeast &1484 &-- &8 &10 &--  &{38.3}  &{38.1}  &{38.0}  &{38.1}  \\
\hline
\end{tabular} 

%% file: tableKulis.tex
\begin{tabular}{|l|c|c|c|c|c|c|c|c|c|}
\hline
&  &  & &  &  & Implicit  & CB  & Implicit  & CB  \\
\raisebox{1.3ex}{Dataset}  &\raisebox{1.3ex}{$N_{\text{train}}$} &\raisebox{1.3ex}{$N_{\text{test}}$} &\raisebox{1.3ex}{$F$} &\raisebox{1.3ex}{$K$} &\raisebox{1.3ex}{RF} &Online &Online &Offline &Offline\\
\hline
breast-cancer & 683 & -- & 9 & 2 & 3.1 & { 3.1 }  & { 3 }  & { 3.1 }  & { 3 }  \\
sonar & 208 & -- & 60 & 2 & 15.1 & { 15.2 }  & { 15.3 }  & { 15.1 }  & { 14.6 }  \\
vowel & 990 & -- & 10 & 11 & 3.2 & { 3.2 }  & { 3.2 }  & { 3.2 }  & { 2.9 } \textbullet\textasteriskcentered \\
ecoli & 336 & -- & 7 & 8 & 13.7 & { 13.7 }  & { 13.6 }  & { 13.7 }  & { 13.6 }  \\
german & 1000 & -- & 24 & 2 & 23.6 & { 23.5 }  & { 23.5 }  & { 23.5 }  & { 23.4 }  \\
glass & 214 & -- & 9 & 6 & 21.4 & { 21.4 }  & { 21.3 }  & { 21.4 }  & { 21 }  \\
image & 2310 & -- & 19 & 7 & 1.9 & { 1.9 }  & { 1.9 }  & { 1.9 }  & { 1.8 } \textbullet \\
ionosphere & 351 & -- & 34 & 2 & 6.4 & { 6.5 }  & { 6.5 }  & { 6.5 }  & { 6.5 }  \\
letter-recognition & 20000 & -- & 16 & 26 & 3.3 & { 3.3 }  & { 3.3 } \textbullet\textasteriskcentered & { 3.3 }  & { 3.3 }  \\
liver-disorders & 345 & -- & 6 & 2 & 26.4 & { 26.4 }  & { 26.4 }  & { 26.4 }  & { 26.4 }  \\
pima-diabetes & 768 & -- & 8 & 2 & 23.2 & { 23.2 }  & { 23.2 }  & { 23.2 }  & { 23.2 }  \\
satimage & 4435 & 2000 & 36 & 6 & 8.8 & { 8.8 }  & { 8.8 }  & { 8.8 }  & { 8.7 } \textbullet \\
vehicle & 846 & -- & 18 & 4 & 24.8 & { 24.7 }  & { 24.9 }  & { 24.7 }  & { 24.9 }  \\
voting-records & 232 & -- & 16 & 2 & 3.5 & { 3.5 }  & { 3.5 }  & { 3.5 }  & { 3.5 }  \\
zipcode & 7291 & 2007 & 256 & 10 & 6.1 & { 6.1 }  & { 6.2 }  & { 6.1 }  & { 6.2 }  \\
abalone & 4177 & -- & 8 & 3 & 45.5 & { 45.5 }  & { 45.6 } \dag & { 45.5 }  & { 45.5 }  \\
balance-scale & 625 & -- & 4 & 3 & 17.7 & { 17.7 }  & { 17.7 }  & { 17.7 }  & { 17.7 }  \\
car & 1728 & -- & 6 & 4 & 2.3 & { 2.3 }  & { 1.8 } \textbullet\textasteriskcentered & { 2.3 }  & { 1.1 } \textbullet\textasteriskcentered \\
connect-4 & 67557 & -- & 42 & 3 & 19.9 & { 19.9 } \textbullet & { 19.5 } \textbullet\textasteriskcentered & { 19.9 } \textbullet & { 18.2 } \textbullet\textasteriskcentered \\
cylinder-bands & 277 & -- & 33 & 2 & 21.4 & { 21.3 }  & { 21.2 }  & { 21.3 }  & { 20.8 } \textbullet \\
hill-valley & 606 & 606 & 100 & 2 & 43.8 & { 43.7 }  & { 43.7 }  & { 43.7 }  & { 43.7 }  \\
isolet & 1559 & -- & 617 & 26 & 6.9 & { 6.9 }  & { 6.9 }  & { 6.9 }  & { 6.9 }  \\
king-rk-vs-king & 28056 & -- & 6 & 18 & 21.6 & { 21.6 } \textbullet & { 19.6 } \textbullet\textasteriskcentered & { 21.5 } \textbullet & { 15.7 } \textbullet\textasteriskcentered \\
king-rk-vs-k-pawn & 3196 & -- & 36 & 2 & 1 & { 1 }  & { 0.7 } \textbullet\textasteriskcentered & { 1 }  & { 0.5 } \textbullet\textasteriskcentered \\
magic & 19020 & -- & 10 & 2 & 11.9 & { 11.9 } \textbullet & { 11.8 } \textbullet\textasteriskcentered & { 11.9 } \textbullet & { 11.7 } \textbullet\textasteriskcentered \\
madelon & 2000 & -- & 500 & 2 & 26.8 & { 26.5 } \textbullet & { 25.6 } \textbullet\textasteriskcentered & { 26.4 } \textbullet & { 21.6 } \textbullet\textasteriskcentered \\
musk & 6598 & -- & 166 & 2 & 1.7 & { 1.7 } \textbullet & { 1.6 } \textbullet\textasteriskcentered & { 1.7 } \textbullet & { 1 } \textbullet\textasteriskcentered \\
splice-junction-gene & 3190 & -- & 59 & 3 & 4.3 & { 4.3 }  & { 4.2 } \textbullet\textasteriskcentered & { 4.3 }  & { 4.1 } \textbullet\textasteriskcentered \\
SAheart & 462 & -- & 9 & 2 & 31.5 & { 31.5 }  & { 31.6 }  & { 31.5 }  & { 31.6 }  \\
yeast & 1484 & -- & 8 & 10 & 37.3 & { 37.3 }  & { 37.3 }  & { 37.3 }  & { 37.3 }  \\
\hline
\end{tabular}